\documentclass[11pt]{article}

\usepackage[utf8]{inputenc}
\usepackage[T1]{fontenc}

\usepackage{times}
\usepackage{microtype}
\usepackage{authblk}

\usepackage{amsmath}
\usepackage{amssymb}

\usepackage{graphicx}
\usepackage{booktabs}
\usepackage{multirow}
\usepackage{array}
\usepackage{tabularx}
\usepackage{caption}
\usepackage{float}
\usepackage{longtable}
\usepackage{subcaption}
\newcolumntype{Y}{>{\raggedright\arraybackslash}X}

\usepackage{xcolor}
\usepackage{url}
\usepackage{hyperref}

\hypersetup{
    colorlinks=true,
    linkcolor=blue,
    citecolor=blue,
    urlcolor=blue
}

\title{
DWT-Fusion: A Signal-Based Framework for Training-Free LLM-Generated Text Detection
}

\author[1]{Mehmet Batuhan \"Ozda\c{s}\thanks{Corresponding author: \texttt{mbozdas@ankara.edu.tr}}}
\author[2]{Murat Osmano\u{g}lu}

\affil[1]{Cyber Security Vocational School, Ankara University, Ankara 06830, T\"urkiye}
\affil[2]{Department of Computer Engineering, Ankara University, Ankara 06830, T\"urkiye}

\date{}

\begin{document}

\maketitle

% -------------------------------------------------
% Abstract
% -------------------------------------------------

\begin{abstract}
Detecting LLM-generated text remains challenging under zero-shot and training-free conditions, especially when detectors must generalize across datasets, domains, and unseen generators. While existing training-free approaches exploit language-model statistics as detection signals, they typically characterize a text through global measures that summarize overall model behavior. Consequently, potentially informative local and multiscale variations in token-level predictability may remain underutilized. Motivated by this observation, we introduce DWT-Fusion, a training-free signal-based framework for detecting LLM-generated text using discrete wavelet analysis of token-level log-probability sequences produced by a proxy causal language model. The proposed framework analyzes these sequences through wavelet-based multiresolution signal representations and derives detection signals from localized probability dynamics. We further evaluate four training-free voting variants, including equal-weight hard voting, equal-weight soft voting, calibration-weighted hard voting, and calibration-weighted soft voting, to combine multiple wavelet configurations without training a supervised meta-classifier. We evaluate the framework on HC3, M4, and MAGE using GPT-Neo-2.7B, GPT-J-6B, Falcon-7B, and LLaMA-3-8B as proxy models. The best single wavelet configurations achieve AUROC values of 0.9872, 0.8185, and 0.7138 on HC3, M4, and MAGE, respectively. With calibration-weighted voting, the best ensemble variants further improve AUROC to 0.9919, 0.8477, and 0.7471. These findings show that DWT-based multiresolution scoring and calibration-guided voting fusion provide effective and interpretable signals for training-free LLM-generated text detection.
\end{abstract}

\noindent\textbf{Keywords:}
LLM-generated text detection; training-free detection; discrete wavelet transform; token log-probability; voting ensemble; zero-shot detection

% -------------------------------------------------
% 1. Introduction
% -------------------------------------------------
\section{Introduction}

Large language models (LLMs) can generate fluent, coherent, and contextually appropriate texts across a wide range of domains. As these systems become increasingly accessible, distinguishing LLM-generated texts from human-written texts has become an important problem for education, academic integrity, online information quality, digital forensics, and content moderation~\cite{fraser2025detecting,wu2024survey}. This detection problem is difficult because generated texts can closely resemble human writing while also varying across domains, prompts, languages, and generator models. The challenge is further amplified by the rapid evolution of LLMs and by the possibility that a text may be produced by a generator that was not available when a detector was developed~\cite{yang2023survey}. Therefore, an effective detector should not only perform well on a fixed dataset, but also provide a detection signal that is less dependent on task-specific training data, source generators, and writing domains.

However, many existing detectors formulate LLM-generated text detection as a supervised classification problem~\cite{wu2024survey}. In this setting, a detector is trained on labeled human-written and machine-generated examples, often using neural encoders, statistical features, learned embeddings, or fine-tuned language models. Such approaches can achieve strong in-domain performance when representative labeled data are available. However, their performance may depend on the training distribution, the source generators, the writing domain, and the availability of reliable labels. This limitation motivates zero-shot and training-free alternatives, where the detector does not learn a task-specific classifier and instead derives evidence directly from the behavior of a pre-trained language model.

Zero-shot detectors commonly use intrinsic statistics obtained from a proxy or source language model. Typical examples include log-likelihood, token rank, log-rank, entropy, likelihood-rank ratio, and probability-curvature-based scores~\cite{mitchell2023detectgpt,su2023detectllm,bao2024fastdetectgpt}. These methods are well-suited to zero-shot detection because they avoid supervised detector training and can be applied directly to unseen texts. However, many of them summarize token-level behavior into a small number of global scalar statistics. While such summaries are efficient and interpretable, they may overlook the sequential structure of token-level predictability. In other words, the order, local fluctuations, and scale-dependent variations of token log-probabilities may contain information that is not fully captured by global likelihood or rank averages.

A more recent direction treats token-level model statistics as signals rather than unordered collections of scalar values. In particular, signal-based approaches show that the token log-probability sequence can be interpreted as a one-dimensional signal and analyzed to capture sequential patterns in model behavior~\cite{luo2026specdetect}. This perspective is important because it preserves the order of token-level predictability values instead of reducing them immediately to global averages. However, preserving the sequence alone does not necessarily capture where fluctuations occur, how local they are, or whether they appear at different temporal scales. As a result, signal representations that mainly summarize the sequence globally may be less sensitive to localized variations that occur only in specific parts of a text or at specific scales. This motivates a representation that can jointly capture local and scale-dependent changes in token predictability.

In this paper, we propose DWT-Fusion, a training-free signal-based framework for LLM-generated text detection. The proposed framework addresses the limitations of global signal summaries by explicitly modeling where token-level probability fluctuations occur and how they vary across different temporal scales. Briefly, the method first extracts token-level conditional log-probabilities from a proxy causal language model and treats the resulting sequence as a one-dimensional signal. The signal is mean-centered to remove global offset effects, then decomposed using the discrete wavelet transform (DWT). Instead of summarizing the probability signal only through global statistics or training a classifier on spectral representations, DWT-Fusion decomposes the mean-centered signal into approximation and detail coefficients across multiple resolutions and derives scalar detection scores from the resulting detail coefficients. This makes it possible to derive detection scores from localized multiresolution energy patterns in the token log-probability signal. In addition to single wavelet-domain scores, we also evaluate voting-based fusion variants that combine multiple wavelet configurations while preserving the training-free nature of the framework. These variants include equal-weight hard voting, equal-weight soft voting, calibration-weighted hard voting, and calibration-weighted soft voting. The method does not train a supervised classifier, fine-tune a language model, construct a reference database, or learn a supervised meta-classifier.

We evaluate the proposed framework on three benchmark datasets for LLM-generated text detection: HC3~\cite{guo2023chatgpt}, M4~\cite{wang2024m4}, and MAGE~\cite{li2024mage}. Token log-probability sequences are extracted using four proxy language models: GPT-Neo-2.7B, GPT-J-6B, Falcon-7B, and LLaMA-3-8B. We compare the proposed wavelet-based scores and voting-based ensemble variants against several zero-shot statistical and spectral baselines, including log-likelihood, rank, log-rank, entropy, LRR, and a DFT-based spectral energy baseline. The results show that the best single wavelet scores are competitive with strong zero-shot baselines on HC3 and achieve higher AUROC than the evaluated statistical and spectral baselines on M4 and MAGE. Furthermore, calibration-weighted voting improves the best held-out AUROC values to 0.9919, 0.8477, and 0.7471 on HC3, M4, and MAGE, respectively. These findings suggest that localized multiresolution variations in token log-probability signals provide complementary evidence for training-free detection, and that calibration-guided voting can further strengthen threshold-independent performance without requiring supervised detector training.

The main contributions of this paper are summarized as follows:

\begin{itemize}
    \item We propose DWT-Fusion, a training-free DWT-based signal scoring framework for LLM-generated text detection using token-level conditional log-probability sequences.

    \item We define three interpretable wavelet-domain scalar scores, namely first-level detail energy, multilevel detail energy, and window-energy variability, to capture localized and multiresolution probability fluctuations.

    \item We introduce calibration-guided voting variants for combining multiple wavelet configurations, including equal-weight hard voting, equal-weight soft voting, calibration-weighted hard voting, and calibration-weighted soft voting, without training a supervised meta-classifier.

    \item We evaluate the proposed framework under a unified protocol across three benchmark datasets, four proxy language models, five wavelet families, and three wavelet-domain score definitions.

    \item We provide a systematic zero-shot evaluation against probability-based, rank-based, entropy-based, likelihood-rank-ratio-based, and DFT-based spectral baselines, allowing the proposed wavelet scores and ensemble variants to be assessed against both statistical and signal-based alternatives.

    \item We show that the best single wavelet configurations achieve held-out AUROC values of 0.9872, 0.8185, and 0.7138 on HC3, M4, and MAGE, respectively, while calibration-weighted voting further improves these values to 0.9919, 0.8477, and 0.7471.
\end{itemize}

\section{Related Work}

In this section, we review representative studies on LLM-generated text detection and organize them according to their methodological assumptions, including zero-shot statistical scoring, intrinsic signal-based detection, supervised classification, feature-based learning, few-shot adaptation, and signal-processing-based approaches. This organization allows us to position DWT-Fusion with respect to both conventional probability-based detectors and more recent methods that preserve token-level sequential information.

Probability- and rank-based zero-shot methods form an important line of LLM-generated text detection research. Mitchell et al.~\cite{mitchell2023detectgpt} introduced DetectGPT, which detects machine-generated text by comparing the log-probability of an original passage with perturbed variants and by exploiting probability curvature in the model distribution. Su et al.~\cite{su2023detectllm} proposed DetectLLM, which leverages token-rank information through LRR and NPR scores to combine likelihood- and rank-based evidence. Bao et al.~\cite{bao2024fastdetectgpt} introduced Fast-DetectGPT to improve the efficiency of curvature-based detection by estimating conditional probability curvature without the expensive perturbation process used in DetectGPT. Yang et al.~\cite{yang2024dnagpt} proposed DNA-GPT, a training-free regeneration-based method that truncates an input text, regenerates continuations, and compares the original and regenerated parts through divergent n-gram analysis or probability divergence. Together, these studies show that probability, rank, curvature, and regeneration-discrepancy signals can support zero-shot detection without training a task-specific detector.

Recent zero-shot and training-free detectors have explored other forms of intrinsic evidence beyond likelihood and rank. Hans et al.~\cite{hans2024binoculars} proposed Binoculars, which compares perplexity and cross-perplexity computed from two language models and uses their ratio as a detection signal. Xu et al.~\cite{xu2025training} introduced Lastde, which treats the token probability sequence as a time series and combines local diversity-entropy statistics with global likelihood information. Ma et al.~\cite{ma2026nts} proposed NTS, which measures normalized temperature sensitivity by observing how surrogate-model probability statistics change under different decoding temperatures. Wu et al.~\cite{wu2025gecscore} introduced GECSCORE, which uses the similarity between an input text and its grammatically corrected version as a zero-shot detection signal. Yang et al.~\cite{yang2025siltd} proposed SILTD, an unsupervised structural detection method based on structural information derived from generative statistics. Sun and Lv~\cite{sun2025textreorder} examined text reordering as a black-box zero-shot signal by comparing the original text with a dependency-based reordered version. These studies demonstrate that training-free detection can be formulated using diverse intrinsic signals, including probability-sequence statistics, temperature sensitivity, grammatical correction, structural information, and text transformation.

Another line of work formulates LLM-generated text detection as a supervised, fine-tuned, or classifier-based problem. Hu et al.~\cite{hu2023radar} proposed RADAR, which trains a detector through adversarial learning against a paraphraser to improve robustness against rewriting attacks. Fu et al.~\cite{fu2025fdllm} introduced FDLLM for black-box LLM fingerprinting, focusing on the identification of source-model characteristics. Verma et al.~\cite{verma2024ghostbuster} proposed Ghostbuster, which extracts probability-based features from weaker language models, searches over feature combinations, and trains a classifier for detecting text generated by black-box or unknown models. Hao et al.~\cite{hao2025learning2rewrite} introduced Learning2Rewrite, which fine-tunes a language model to amplify rewriting discrepancies between human-written and LLM-generated texts. Zeng et al.~\cite{zeng2025humanoutliers} proposed Human Texts Are Outliers, which formulates detection through an outlier-based learning perspective. Zheng et al.~\cite{zheng2025lm2otifs} introduced LM2OTIFS, which approaches detection through graph-based representation learning.

Feature-based classifiers further expand this supervised detection line. Titze and Halvani~\cite{titze2025logaid} proposed LOG-AID, which extracts logit-based statistical features such as surprisal, entropy, divergence, and log-rank before applying logistic regression. Cheng et al.~\cite{cheng2024biscope} introduced BISCOPE, which uses bidirectional cross-entropy statistics to measure how strongly a model memorizes preceding and following token contexts. Wang et al.~\cite{wang2025benatten} proposed BENATTEN, which analyzes attention distributions using Benford's law-inspired features and trains neural classifiers on the resulting representations. Wang et al.~\cite{wang2026la2hdetect} proposed LA2HDetect, which extracts latent-level language features such as complexity, potentiality, and logicality and uses ensemble learning for detection. Le and Tran~\cite{le2025metricdet} developed a metric-based detection system that maps texts into embedding spaces and learns distance-based separation between human-written and machine-generated texts. These methods enrich the detection feature space, but their final decisions depend on trained classifiers, learned ensembles, or task-specific decision functions.

Few-shot and adaptation-based detectors aim to improve generalization under limited supervision, domain shift, or evolving model sources. Bao et al.~\cite{bao2025hart2d} proposed a HART-style two-dimensional detection framework that decouples content and expression and fits a binary decision boundary using a small development set. Chen et al.~\cite{chen2025divscore} introduced DivScore for specialized domains such as medical and legal text detection by using domain adaptation and entropy-based scoring. Guo et al.~\cite{guo2024detective} proposed DeTeCtive, which uses multi-level contrastive learning and retrieval mechanisms to improve out-of-distribution generalization. He et al.~\cite{he2025detree} introduced DETree for detecting human--AI collaborative texts through tree-structured hierarchical representation learning. Zhou et al.~\cite{zhou2025adadetectgpt} proposed AdaDetectGPT, which adaptively learns a witness function over token-level statistics with statistical guarantees. Li and Wang~\cite{li2025continualorigin} studied continual origin tracing by constructing prototype-based representations for evolving LLM sources. These studies address important practical issues such as domain adaptation, robustness, and evolving source attribution.

The closest line of work to this paper is signal-based LLM-generated text detection. Instead of treating token-level model statistics as unordered aggregate values, signal-based methods preserve their sequential structure. Luo et al.~\cite{luo2026specdetect} proposed SpecDetect, which represents token log-probabilities as a one-dimensional signal and applies the discrete Fourier transform (DFT) to compute spectral energy as a training-free detection score. This work shows that token log-probability dynamics contain useful signal-level evidence, but DFT-based analysis primarily captures global spectral characteristics. Liu et al.~\cite{liu2026wavedetect} introduced WAVEDETECT, a supervised wavelet-based framework that converts token probability sequences into continuous wavelet transform (CWT) time-frequency representations and trains a CNN-based spectral classifier. These studies are directly relevant because they treat token-level probability behavior as a signal and analyze its spectral or wavelet-domain structure.

Table~\ref{tab:related_positioning} provides a methodological comparison of the representative detection methods discussed in this section. In the table, TF denotes whether the method avoids supervised detector training or fine-tuning, token-seq. indicates whether token-level probability, log-probability, or perplexity sequences are explicitly used as ordered signals, and local/multi. indicates whether the method explicitly models local or multiscale signal structure.

{\tiny
\setlength{\tabcolsep}{1.5pt}
\renewcommand{\arraystretch}{1.06}

\begin{longtable}{@{}
>{\raggedright\arraybackslash}p{1.65cm}
>{\raggedright\arraybackslash}p{1.95cm}
>{\raggedright\arraybackslash}p{2.15cm}
>{\centering\arraybackslash}p{0.75cm}
>{\centering\arraybackslash}p{0.95cm}
>{\centering\arraybackslash}p{0.55cm}
>{\raggedright\arraybackslash}p{4.10cm}
@{}}
\caption{Methodological comparison of LLM-generated text detection studies discussed in the related work.}
\label{tab:related_positioning}\\
\toprule
\textbf{Method} & \textbf{Paradigm} & \textbf{Main signal} & \textbf{Token-seq.?} & \textbf{Local / multi.?} & \textbf{TF?} & \textbf{Limitation} \\
\midrule
\endfirsthead

\toprule
\textbf{Method} & \textbf{Paradigm} & \textbf{Main signal} & \textbf{Token-seq.?} & \textbf{Local / multi.?} & \textbf{TF?} & \textbf{Limitation} \\
\midrule
\endhead

DetectGPT~\cite{mitchell2023detectgpt}
& Zero-shot, probability-based
& Probability curvature
& No
& No
& Yes
& Relies on perturbation-based curvature estimation and summarizes behavior globally. \\

DetectLLM~\cite{su2023detectllm}
& Zero-shot, rank-based
& Likelihood, rank, LRR, NPR
& No
& No
& Yes
& Relies on global likelihood and rank aggregation without modeling local sequence structure. \\

Fast-DetectGPT~\cite{bao2024fastdetectgpt}
& Zero-shot, probability-based
& Conditional probability curvature
& No
& No
& Yes
& Improves efficiency but still uses global curvature-based scoring. \\

DNA-GPT~\cite{yang2024dnagpt}
& Training-free, regeneration-based
& N-gram divergence and probability divergence
& No
& No
& Yes
& Requires truncation and regeneration to estimate text-level discrepancy. \\

Binoculars~\cite{hans2024binoculars}
& Zero-shot, perplexity-based
& Perplexity and cross-perplexity ratio
& No
& No
& Yes
& Requires two language models and relies on global perplexity comparison. \\

Lastde~\cite{xu2025training}
& Training-free, sequence-based
& Local diversity entropy and likelihood
& Yes
& Yes
& Yes
& Captures local/multiscale structure through time-domain statistics rather than wavelet-domain decomposition. \\

NTS~\cite{ma2026nts}
& Zero-shot, temperature-based
& Normalized temperature sensitivity
& No
& No
& Yes
& Requires temperature-dependent probability probing and does not model local multiscale structure. \\

GECSCORE~\cite{wu2025gecscore}
& Zero-shot, correction-based
& Correction similarity
& No
& No
& Yes
& Depends on correction-induced differences rather than the original token-probability signal. \\

SILTD~\cite{yang2025siltd}
& Unsupervised, structural
& Structural information
& No
& No
& Yes
& Relies on structural modeling rather than explicit token-level signal decomposition. \\

Text Reordering~\cite{sun2025textreorder}
& Zero-shot, transformation-based
& Reordering-based discrepancy
& No
& No
& Yes
& Requires text transformation and comparison with dependency-based reordered variants. \\

RADAR~\cite{hu2023radar}
& Supervised, adversarial
& Adversarial rewriting signal
& No
& No
& No
& Requires supervised detector--paraphraser adversarial training. \\

FDLLM~\cite{fu2025fdllm}
& Supervised, fingerprinting
& Source-model fingerprints
& No
& No
& No
& Focuses on source-model fingerprinting rather than training-free human--LLM text separation. \\

Ghostbuster~\cite{verma2024ghostbuster}
& Supervised, black-box
& Probability-based searched features
& No
& No
& No
& Requires feature search and supervised classifier training over weaker-LM features. \\

Learning2 Rewrite~\cite{hao2025learning2rewrite}
& Fine-tuned, rewriting-based
& Rewriting discrepancy
& No
& No
& No
& Requires fine-tuning a rewriting model to amplify human--machine discrepancies. \\

Human Texts Are Outliers~\cite{zeng2025humanoutliers}
& Learning-based detection
& Human-text outlier evidence
& No
& No
& No
& Depends on a learning-based outlier formulation rather than training-free scalar scoring. \\

LM2OTIFS~\cite{zheng2025lm2otifs}
& Graph-based detection
& Learned graph representations
& No
& No
& No
& Depends on graph-based learned representations rather than direct signal-domain scoring. \\

LOG-AID~\cite{titze2025logaid}
& Feature-based classifier
& Surprisal, entropy, divergence, log-rank
& No
& No
& No
& Requires a logistic regression classifier over logit-based statistical features. \\

BISCOPE~\cite{cheng2024biscope}
& Feature-based classifier
& Bidirectional cross-entropy
& No
& No
& No
& Relies on classifier-based use of bidirectional cross-entropy features. \\

BENATTEN~\cite{wang2025benatten}
& Neural classifier
& Benford-inspired attention statistics
& No
& No
& No
& Requires neural classifier training on attention-distribution features. \\

LA2HDetect~\cite{wang2026la2hdetect}
& Ensemble learning
& Complexity, potentiality, logicality
& No
& No
& No
& Uses a learned ensemble over latent-level language features. \\

Metric-based Detection~\cite{le2025metricdet}
& Metric learning
& Text embedding distances
& No
& No
& No
& Requires metric learning over embedding-space representations. \\

HART / 2D Detection~\cite{bao2025hart2d}
& Few-shot, 2D detection
& Content and expression dimensions
& No
& No
& No
& Requires a development set to fit a decision boundary. \\

DivScore~\cite{chen2025divscore}
& Domain-adaptive detection
& Entropy and domain-aware signal
& No
& No
& No
& Relies on domain adaptation for specialized detection settings. \\

DeTeCtive~\cite{guo2024detective}
& Contrastive detection
& Multi-level representations
& No
& No
& No
& Requires contrastive representation learning and retrieval mechanisms. \\

DETree~\cite{he2025detree}
& Hierarchical representation learning
& Hierarchical representations
& No
& No
& No
& Relies on hierarchical representation learning rather than training-free signal scoring. \\

AdaDetectGPT~\cite{zhou2025adadetectgpt}
& Adaptive statistical detection
& Learned witness function
& No
& No
& No
& Learns an adaptive witness function over token-level statistics. \\

Continual Origin Tracing~\cite{li2025continualorigin}
& Continual source tracing
& Prototype-based origin representation
& No
& No
& No
& Targets evolving source attribution and relies on prototype-based representations. \\

SpecDetect~\cite{luo2026specdetect}
& Training-free, spectral signal-based
& DFT spectral energy
& Yes
& No
& Yes
& Captures global Fourier-domain energy without localized multiresolution analysis. \\

WaveDetect~\cite{liu2026wavedetect}
& Supervised, wavelet-based
& CWT time-frequency representation
& Yes
& Yes
& No
& Requires supervised CNN training over CWT time-frequency representations. \\

\bottomrule
\end{longtable}
}

Although prior work has made substantial progress in LLM-generated text detection, Table~\ref{tab:related_positioning} shows that the design space remains unevenly covered. Classical zero-shot detectors such as DetectGPT, DetectLLM, Fast-DetectGPT, DNA-GPT, Binoculars, NTS, GECSCORE, SILTD, and text-reordering methods avoid supervised detector training, but they generally operate through global probability, rank, curvature, regeneration, correction, structural, or transformation-based signals rather than localized multiresolution signal analysis. Supervised and learning-based methods, including RADAR, FDLLM, Ghostbuster, Learning2Rewrite, LOG-AID, BISCOPE, BENATTEN, LA2HDetect, metric-learning approaches, and graph- or representation-based detectors, enrich the detection feature space and address practical robustness or generalization issues, but their final decisions depend on trained classifiers, fine-tuned models, learned ensembles, or task-specific decision functions. Few-shot and adaptation-based methods further improve generalization under limited supervision or domain shift, but they also rely on development data, adaptation mechanisms, retrieval, learned witness functions, or prototype-based representations.

Among sequence-aware and signal-processing-based methods, Lastde is particularly relevant because it treats token probability sequences as time-domain evidence and combines local diversity entropy with likelihood information. SpecDetect further demonstrates that token log-probability sequences can be treated as one-dimensional signals, but it relies on global Fourier-domain spectral energy and therefore does not explicitly model localized multiresolution variations. WAVEDETECT is the closest wavelet-based natural-language detector because it converts token probability sequences into CWT time-frequency representations; however, it trains a CNN-based spectral classifier and therefore belongs to a supervised setting. In contrast, DWT-Fusion combines three properties that are not jointly covered by these prior methods: it remains training-free, directly analyzes the token log-probability sequence as an ordered signal, and derives scalar scores from localized multiresolution DWT energy patterns without training a classifier, fine-tuning a proxy model, or learning a supervised meta-classifier. The calibration-guided hard and soft voting variants further allow multiple wavelet configurations to be combined at the score level while preserving this training-free formulation. Thus, the proposed framework fills the gap between global zero-shot statistics, time-domain probability-sequence statistics, global Fourier-domain scoring, and supervised wavelet-based classification.

% -------------------------------------------------
% 3. Methodology
% -------------------------------------------------

\section{Methodology}

This section describes DWT-Fusion, the proposed DWT-based signal scoring and voting-fusion framework for training-free LLM-generated text detection. The central idea is to represent each input text as an ordered token log-probability signal and to analyze this signal in the wavelet domain. Unlike conventional zero-shot detectors that summarize a text using global likelihood, rank, or entropy statistics, the proposed method preserves the sequential structure of token-level probabilities and derives localized multiresolution scores from this sequence. In addition to using individual wavelet-domain scores, the framework also evaluates training-free voting ensembles that combine multiple wavelet configurations without training a supervised meta-classifier.

\begin{figure}[!t]
    \centering
    \includegraphics[width=\linewidth]{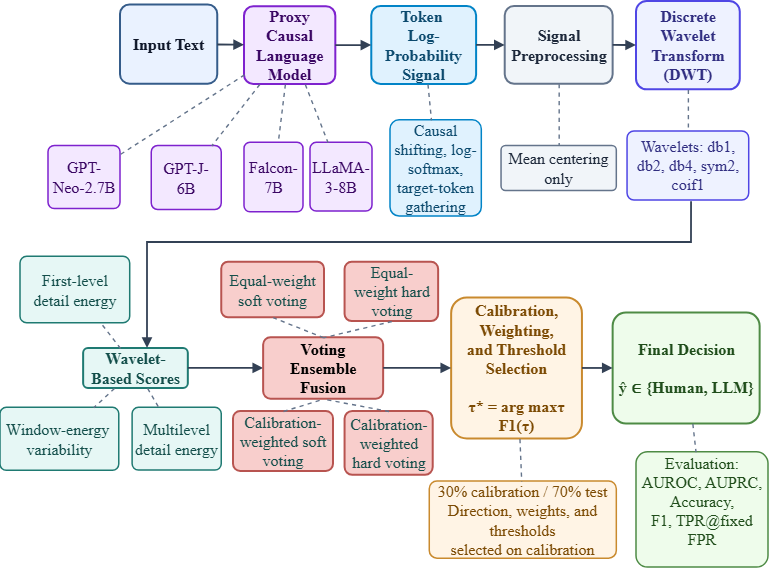}
    \caption{Overview of DWT-Fusion, the proposed DWT-based signal scoring and calibration-guided voting framework for training-free LLM-generated text detection.}
    \label{fig:pipeline}
\end{figure}

As shown in Figure~\ref{fig:pipeline}, the proposed framework consists of six main stages: (i) token log-probability signal extraction, where a proxy causal language model assigns conditional log-probabilities to the observed token sequence; (ii) signal preprocessing, where the resulting one-dimensional signal is mean-centered to remove global offset effects; (iii) discrete wavelet decomposition, where the preprocessed signal is decomposed into approximation and detail coefficients across multiple resolutions; (iv) wavelet-based score computation, where scalar detection scores are derived from localized energy patterns in the detail coefficients; (v) voting-based ensemble fusion, where multiple wavelet configurations are combined using equal-weight or calibration-weighted hard and soft voting; and (vi) calibration-based threshold selection and final decision making, where score direction, ensemble weights, and thresholds are determined on the calibration split and then applied to the held-out test split.

\subsection{Token Log-Probability Signal Extraction}

Let $X=(t_1,t_2,\ldots,t_n)$ denote an input text represented as a sequence of tokens under a proxy causal language model $p_\theta$. For each token position, we compute the conditional log-probability assigned by the proxy model to the observed next token. Formally, the raw token log-probability signal is defined as

\begin{equation}
x_i = \log p_\theta(t_{i+1} \mid t_{\leq i}), \qquad i=1,\ldots,n-1,
\end{equation}

where $x_i$ denotes the conditional log-probability assigned to the next observed token $t_{i+1}$ given the preceding prefix $t_{\leq i}$.

In implementation, each input text is tokenized with a maximum sequence length of 512 tokens. This fixed limit standardizes signal extraction across proxy language models and reduces GPU memory and inference-time requirements during large-scale evaluation. Longer texts are truncated to this limit before token log-probability computation. The tokenized sequence is passed through the proxy causal language model to obtain the output logits. Since causal language models predict the next token from the previous context, we align the logits and target tokens using the standard shifted formulation: logits at position $i$ are paired with the target token at position $i+1$. A log-softmax operation is applied over the vocabulary dimension, and the log-probability corresponding to the observed target token is selected. Due to this causal shifting, the first token does not contribute a conditional log-probability value to the final signal. For batch processing, padding tokens are excluded using the attention mask.

The resulting sequence of valid target-token log-probabilities is converted into a one-dimensional numerical signal:

\begin{equation}
\mathbf{x} = (x_1,x_2,\ldots,x_N),
\end{equation}

where $N$ denotes the number of valid conditional log-probability values after truncation, shifting, and padding-mask removal.

\subsection{Signal Preprocessing}

Before applying the discrete wavelet transform, we use a minimal preprocessing strategy consisting of two steps: a validity check for extremely short sequences and mean-centering of the token log-probability signal. First, the extracted log-probability sequence is converted into a numerical array. Then, sequences with fewer than four valid log-probability values are excluded from the wavelet scoring stage. Such short sequences contain too few samples to provide a meaningful local fluctuation pattern, and their wavelet coefficients would be strongly affected by boundary effects rather than by reliable multiresolution structure. This length check only removes degenerate cases. The actual decomposition level is determined later for each valid sequence according to the signal length and the wavelet filter length.

After this validity check, each remaining sequence is mean-centered by subtracting its global arithmetic mean:

\begin{equation}
\mu_x = \frac{1}{N}\sum_{i=1}^{N} x_i,
\end{equation}

\begin{equation}
\tilde{x}_i = x_i - \mu_x.
\end{equation}

Mean-centering removes the global offset of the token log-probability signal before wavelet analysis. This step is important because the proposed detector focuses on local and scale-dependent fluctuations rather than only on the overall likelihood level of a text. By subtracting the sequence mean, the DWT coefficients emphasize variations around each text's own average predictability level. As a result, the resulting wavelet-domain scores are more directly tied to the structure of token-level fluctuations.

The final signal used for wavelet analysis is therefore

\begin{equation}
\tilde{\mathbf{x}} = (\tilde{x}_1,\tilde{x}_2,\ldots,\tilde{x}_N).
\end{equation}

Apart from the short-sequence validity check and mean-centering, no additional normalization, clipping, or smoothing is applied before the wavelet transform. This choice preserves the relative magnitude of local fluctuations in the proxy-model predictability signal while removing only the global offset component.

\subsection{Discrete Wavelet Decomposition}

After preprocessing, we apply a multilevel discrete wavelet transform (DWT) to the mean-centered token log-probability signal $\tilde{\mathbf{x}}$ following the standard multiresolution wavelet decomposition framework~\cite{mallat1989wavelet, daubechies1992wavelets}. In this stage, the signal is decomposed into one approximation component and multiple detail components:

\begin{equation}
\{A_L,D_L,D_{L-1},\ldots,D_1\} = \mathrm{DWT}(\tilde{\mathbf{x}}),
\end{equation}

where $A_L$ denotes the approximation coefficients at decomposition level $L$, and $D_l$ denotes the detail coefficients at level $l$. The approximation coefficients represent lower-frequency trends in the signal, whereas the detail coefficients represent localized fluctuations at different resolutions.

The DWT is computed using symmetric boundary extension. Since input texts may produce log-probability signals of different lengths, the decomposition level is selected dynamically for each sequence. Specifically, the selected level is the minimum between a predefined maximum level $L_{\max}=3$ and the maximum level allowed by the signal length and the wavelet filter length:

\begin{equation}
L = \min(L_{\max}, L_{\mathrm{allowed}}).
\end{equation}

We set $L_{\max}=3$ to capture local, intermediate, and coarser-scale fluctuations while avoiding overly deep decompositions that may produce very short detail coefficient sequences for shorter texts. This setting provides a practical balance between multiresolution analysis and score stability across variable-length inputs. If the resulting level is smaller than one, the wavelet decomposition is skipped for that sequence.

\begin{figure}[!t]
    \centering
    \includegraphics[width=\linewidth]{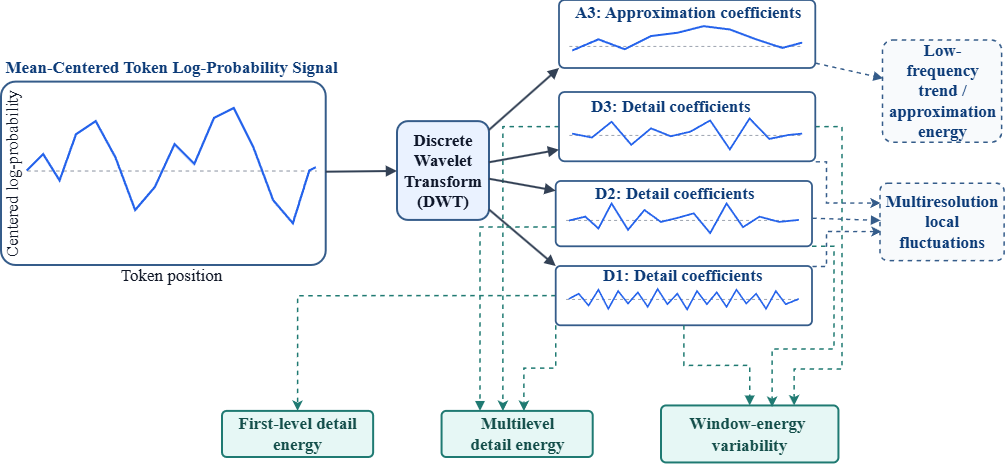}
    \caption{Wavelet-domain representation of the mean-centered token log-probability signal. The DWT is applied using one of the evaluated mother wavelets, $\psi \in \{\mathrm{db1}, \mathrm{db2}, \mathrm{db4}, \mathrm{sym2}, \mathrm{coif1}\}$, and the resulting detail coefficients are used to compute the three scalar wavelet scores: first-level detail energy, multilevel detail energy, and window-energy variability.}
    \label{fig:wavelet_decomposition}
\end{figure}

After decomposition, the detail coefficients are organized by level as $D_1$, $D_2$, and $D_3$ whenever the corresponding levels are available. If a signal is too short to support all three levels, unavailable detail levels are treated as missing and are excluded from score computations that aggregate over valid levels. The method can be applied with different compactly supported mother wavelets; the specific wavelet families evaluated in this study are specified in Section~\ref{sec:experimental_setup}. Figure~\ref{fig:wavelet_decomposition} illustrates how the mean-centered signal is decomposed into approximation and detail components and how the detail coefficients support the three scalar wavelet scores considered in this study.

\subsection{Wavelet-Based Detection Scores}
\label{subsec:wavelet_scores}
After the DWT decomposition, the proposed framework computes three scalar wavelet-domain detection scores from the detail coefficients. These scores should be distinguished from the wavelet families used to perform the decomposition. The score definitions determine how the detail coefficients are summarized into a scalar detection value, whereas the wavelet family determines the basis functions used by the DWT. In the experiments, the same three score definitions are evaluated with different wavelet families.

Let $D_l = (D_{l,1},D_{l,2},\ldots,D_{l,N_l})$ denote the detail coefficients at level $l$. The first score is the first-level normalized detail energy. For the first-level detail coefficients $D_1$, we compute the mean squared coefficient magnitude as

\begin{equation}
E_1 = \frac{1}{N_1}\sum_{i=1}^{N_1} D_{1,i}^{2}.
\end{equation}

This score captures short-scale fluctuations in the mean-centered token log-probability signal and corresponds to the energy\_norm score in the implementation.

The second score is the multilevel detail energy. For each valid detail level $l$, we first compute the normalized detail energy

\begin{equation}
E_l = \frac{1}{N_l}\sum_{i=1}^{N_l} D_{l,i}^{2}.
\end{equation}

The multilevel detail energy is then obtained by summing the normalized energies across the available detail levels up to level 3:

\begin{equation}
E_{\mathrm{multi}} = \sum_{l \in \mathcal{L}} E_l,
\end{equation}

where $\mathcal{L}\subseteq \{1,2,3\}$ denotes the set of valid detail levels available for the input sequence. If a signal is too short to support a particular decomposition level, that level is excluded from the summation. This score captures the overall amount of localized variation in the token log-probability signal across multiple resolutions.

The third score is window-energy variability. This score measures how strongly wavelet energy changes across local regions of the text. For each valid detail level $l$, the detail coefficient sequence is divided into non-overlapping windows of size $w$. Let $e_{l,j}$ denote the mean squared energy of the $j$-th window:

\begin{equation}
e_{l,j} = \frac{1}{w}\sum_{i \in \mathcal{W}_{l,j}} D_{l,i}^{2},
\end{equation}

where $\mathcal{W}_{l,j}$ is the index set of the $j$-th window at level $l$. The variability of local wavelet energy at level $l$ is computed as the standard deviation of the window energies:

\begin{equation}
\sigma_l = \mathrm{std}(e_{l,1}, e_{l,2}, \ldots, e_{l,m_l}),
\end{equation}

where $m_l$ is the number of valid windows at level $l$. The final window-based score is obtained by averaging this variability over the valid detail levels:

\begin{equation}
W_{\mathrm{std}} = \frac{1}{|\mathcal{L}_{w}|}\sum_{l \in \mathcal{L}_{w}} \sigma_l,
\end{equation}

where $\mathcal{L}_{w}$ denotes the set of detail levels with valid window-energy statistics. This score corresponds to \texttt{window\_std} in the implementation and captures whether wavelet-domain energy is distributed uniformly or varies strongly across different local regions of the text.

Thus, the framework uses three scalar wavelet-domain scores: first-level detail energy, multilevel detail energy, and window-energy variability. These scores can be used either individually as training-free detectors or combined through voting-based ensemble variants. The main experimental results report both the best-performing single wavelet configurations and the best-performing ensemble variants for each dataset and operating condition.

\subsection{Calibration-Based Voting Ensembles}
\label{subsec:voting_ensembles}

In addition to evaluating individual wavelet-domain scores, we also investigate voting-based ensemble variants that combine multiple wavelet configurations. Each base configuration corresponds to a specific combination of proxy language model, wavelet family, and wavelet-domain score definition. Let $\mathcal{K}=\{1,2,\ldots,K\}$ denote the set of base configurations included in a given ensemble, and let $s_k(X)$ denote the direction-corrected scalar score produced by configuration $k$ for input text $X$. Score direction is determined only on the calibration split so that larger scores consistently indicate stronger evidence of LLM-generated text.

We evaluate four training-free voting variants: equal-weight hard voting, equal-weight soft voting, calibration-weighted hard voting, and calibration-weighted soft voting. These variants do not train a supervised meta-classifier. The calibration split is used only for score-direction correction, threshold selection, score normalization, and deterministic weight computation.

For hard voting, each base configuration is first converted into a binary prediction using its calibration-selected threshold $\tau_k$:

\begin{equation}
\hat{y}_k(X) =
\begin{cases}
1, & s_k(X) \geq \tau_k, \\
0, & s_k(X) < \tau_k,
\end{cases}
\end{equation}

where $\hat{y}_k(X)=1$ denotes LLM-generated text and $\hat{y}_k(X)=0$ denotes human-written text. In equal-weight hard voting, the ensemble vote score is computed as

\begin{equation}
V_{\mathrm{hard}}(X) = \frac{1}{K}\sum_{k=1}^{K}\hat{y}_k(X).
\end{equation}

The final ensemble prediction is obtained by majority voting:

\begin{equation}
\hat{y}_{\mathrm{ens}}(X) =
\begin{cases}
1, & V_{\mathrm{hard}}(X) \geq 0.5, \\
0, & V_{\mathrm{hard}}(X) < 0.5.
\end{cases}
\end{equation}

For calibration-weighted hard voting, each base configuration is assigned a non-negative deterministic weight $w_k$ computed only from calibration-set performance. The weighted hard-voting score is

\begin{equation}
V_{\mathrm{whard}}(X) =
\frac{\sum_{k=1}^{K} w_k \hat{y}_k(X)}
{\sum_{k=1}^{K} w_k}.
\end{equation}

The final prediction is again obtained by applying a threshold of 0.5 to the weighted vote score.

For soft voting, the continuous wavelet-domain scores are first normalized using calibration-set statistics. For each base configuration $k$, we compute

\begin{equation}
z_k(X) = \frac{s_k(X)-\mu_k^{\mathrm{cal}}}{\sigma_k^{\mathrm{cal}}+\epsilon},
\end{equation}

where $\mu_k^{\mathrm{cal}}$ and $\sigma_k^{\mathrm{cal}}$ denote the calibration-set mean and standard deviation of the direction-corrected scores, and $\epsilon$ is a small numerical constant for stability. The normalized score is then converted into a pseudo-probability using the sigmoid function:

\begin{equation}
p_k(X)=\frac{1}{1+\exp(-z_k(X))}.
\end{equation}

In equal-weight soft voting, the ensemble score is the average pseudo-probability across base configurations:

\begin{equation}
P_{\mathrm{soft}}(X)=\frac{1}{K}\sum_{k=1}^{K}p_k(X).
\end{equation}

In calibration-weighted soft voting, the ensemble score is computed as

\begin{equation}
P_{\mathrm{wsoft}}(X)=
\frac{\sum_{k=1}^{K}w_k p_k(X)}
{\sum_{k=1}^{K}w_k}.
\end{equation}

The final soft-voting prediction is obtained by selecting a threshold on the calibration split and applying it unchanged to the held-out test split.

For the calibration-weighted variants, we evaluate deterministic weighting rules based on calibration-set performance, including AUROC-based weighting, AUPRC-based weighting, TPR@1\%FPR-based weighting, and rank-based weighting. These rules assign larger weights to configurations that perform better on the calibration split. If all weights become zero under a given rule, equal weights are used as a fallback. Since the weights are computed directly from calibration-set summary metrics and no model parameters are optimized, the ensemble variants remain training-free.

The voting ensembles are evaluated at different scopes, including score-level ensembles, wavelet-family ensembles, proxy-model ensembles, and full configuration ensembles. The full configuration ensemble combines all wavelet-based configurations, corresponding to four proxy language models, five wavelet families, and three wavelet-domain scores.

% -------------------------------------------------
% 4. Experimental Setup
% -------------------------------------------------
\section{Experimental Setup}
\label{sec:experimental_setup}

This section describes the experimental configuration used to evaluate the proposed wavelet-based and voting-ensemble framework across different datasets, proxy language models, wavelet families, baseline methods, and ensemble variants. We first present the implementation details, including model inference, wavelet parameters, and voting configurations. We then describe the datasets, proxy language models, baseline scores, ensemble settings, and calibration-test evaluation protocol used throughout the experiments.

\subsection{Implementation Details}

All experiments are implemented in Python and executed in a Google Colab environment with GPU acceleration. We use PyTorch and Hugging Face Transformers for proxy language model inference, PyWavelets for discrete wavelet decomposition, and scikit-learn for threshold selection and evaluation. Data processing and numerical operations are performed using NumPy, pandas, SciPy, and the Hugging Face Datasets library.

To ensure reproducibility, the random seed is fixed to 42 across Python, NumPy, and PyTorch. All input texts are tokenized with a maximum length of 512 tokens. The proxy language models are used only in inference mode; no model parameters are updated. Models are loaded in half precision using torch.float16 with automatic device mapping. Token log-probability extraction is performed in batches, with a batch size of 16 in the implementation.

\begin{table}[!t]
\centering
\caption{Main experimental configuration.}
\label{tab:experimental_config}
\small
\begin{tabularx}{\linewidth}{l>{\raggedright\arraybackslash}X}
\toprule
Configuration & Value \\
\midrule
Environment & Python in Google Colab with GPU acceleration \\
Main libraries & PyTorch, Hugging Face Transformers, PyWavelets, scikit-learn, NumPy, pandas, SciPy \\
Random seed & 42 \\
Maximum input length & 512 tokens \\
Calibration / test split & 30\% / 70\% \\
Proxy model mode & Inference only, no fine-tuning \\
Proxy model precision & \texttt{torch.float16} \\
Batch size for scoring & 16 \\
Wavelet families & \texttt{db1}, \texttt{db2}, \texttt{db4}, \texttt{sym2}, \texttt{coif1} \\
Maximum DWT level & 3 \\
DWT boundary mode & Symmetric \\
Window size for local energy variability & 8 \\
Wavelet-domain scores & \texttt{energy\_norm}, \texttt{multilevel\_energy}, \texttt{window\_std} \\
Voting variants & Equal-weight hard, equal-weight soft, calibration-weighted hard, calibration-weighted soft \\
Weighting schemes & AUROC-based, AUPRC-based, TPR@1\%FPR-based, rank-based \\
Full ensemble size & 60 configurations: 4 proxy models $\times$ 5 wavelet families $\times$ 3 scores \\
\bottomrule
\end{tabularx}
\end{table}

For wavelet-based scoring, we evaluate five compactly supported wavelet families: db1, db2, db4, sym2, and coif1. These wavelets are selected to cover simple, short-support filters as well as smoother and more structured wavelet bases. This allows us to examine whether detection performance depends on the choice of mother wavelet while keeping the search space limited and computationally feasible. The same three scalar wavelet-domain scores are computed for each wavelet family. The maximum DWT decomposition level is set to 3, and the symmetric boundary mode is used. For the window-energy variability score, the window size is set to 8. Table~\ref{tab:experimental_config} summarizes the main experimental configuration used throughout the evaluation.

\subsection{Datasets}

We evaluate the proposed method on three benchmark datasets for LLM-generated text detection: HC3~\cite{guo2023chatgpt}, M4~\cite{wang2024m4}, and MAGE~\cite{li2024mage}. These datasets provide complementary evaluation settings. HC3 represents a human-versus-ChatGPT detection scenario, while M4 and MAGE provide broader and more heterogeneous evaluation conditions involving multiple domains, languages, or generator models. Using these datasets allows us to examine whether wavelet-based token log-probability analysis remains useful beyond a single dataset or generation source.

All datasets are converted into a unified binary format, where human-written texts are assigned label $0$ and LLM-generated texts are assigned label $1$. Texts containing fewer than five words are removed during dataset normalization. Duplicate scored records are removed during the result consolidation stage using the model, dataset, and text identifier fields. Table~\ref{tab:dataset_summary} reports the final dataset sizes used in the evaluation after filtering short texts and consolidating the scored records.

\begin{table}[!t]
\centering
\caption{Summary of the datasets used in the experiments after filtering short texts and consolidating duplicate scored records.}
\label{tab:dataset_summary}
\small
\begin{tabularx}{\linewidth}{lrrr>{\raggedright\arraybackslash}X}
\toprule
Dataset & Total & Human & LLM-Generated & Notes \\
\midrule
HC3  & 48,179  & 24,315  & 23,864  & Human-written and ChatGPT-generated answers \\
M4   & 162,882 & 82,337  & 80,545  & Multi-domain / multilingual benchmark \\
MAGE & 436,606 & 284,224 & 152,382 & Multi-generator heterogeneous benchmark \\
\bottomrule
\end{tabularx}
\end{table}

\subsection{Proxy Language Models}

The proposed framework uses proxy causal language models to extract token-level conditional log-probabilities from each input text. In our experiments, we evaluate four pre-trained causal language models as proxy scorers: GPT-Neo-2.7B, GPT-J-6B, Falcon-7B, and LLaMA-3-8B. These models are used only to compute token-level probability signals; their parameters are not updated, and no task-specific detector is fine-tuned.

Each proxy model is used in inference mode only. We load the models with the Hugging Face AutoTokenizer and AutoModelForCausalLM interfaces, using half precision where supported and automatic device mapping for GPU execution. If a tokenizer does not define a padding token, we use its end-of-sequence token for padding. The same token log-probability extraction procedure is applied across all proxy models to ensure a consistent scoring protocol.

\subsection{Baseline Methods}

We compare the proposed wavelet-based scores with several zero-shot statistical and spectral baselines. The statistical baselines include mean log-likelihood, mean token rank, mean log-rank, mean entropy, and likelihood-rank ratio (LRR). These methods are computed from the proxy model outputs and summarize each text using global token-level probability, rank, or uncertainty statistics.

For a token sequence with valid target positions $i=1,\ldots,N$, the mean log-likelihood baseline is computed as the average conditional log-probability of the observed tokens. The rank baseline computes the average vocabulary rank of the observed tokens under the proxy model distribution, while the log-rank baseline averages the logarithm of these ranks. The entropy baseline averages the full-vocabulary Shannon entropy at each token position. The LRR baseline is computed as the ratio between mean log-likelihood and mean log-rank, with a small numerical constant added to the denominator for stability.

We also include a DFT-based spectral energy baseline inspired by signal-based detection. This baseline applies a discrete Fourier transform to the raw target token log-probability sequence and computes the length-normalized total spectral energy. This comparison is important because it directly evaluates whether localized multiresolution wavelet analysis provides additional value over global Fourier-domain spectral energy.

\subsection{Ensemble Configurations}

In addition to individual wavelet-domain scores, we evaluate four voting-based ensemble variants: equal-weight hard voting, equal-weight soft voting, calibration-weighted hard voting, and calibration-weighted soft voting. These variants combine multiple wavelet configurations without training a supervised meta-classifier. Each base configuration is defined by a proxy language model, a wavelet family, and a wavelet-domain score.

We evaluate ensemble fusion at four scopes. The score-level ensemble combines the three wavelet-domain scores for a fixed dataset, proxy language model, and wavelet family. The wavelet-family ensemble combines the five wavelet families for a fixed dataset, proxy language model, and score definition. The proxy-model ensemble combines the four proxy language models for a fixed dataset, wavelet family, and score definition. Finally, the full configuration ensemble combines all available wavelet-based configurations, corresponding to $4 \times 5 \times 3 = 60$ base detectors.

For equal-weight hard voting, each base configuration first produces a binary decision using its calibration-selected threshold, and the final prediction is obtained by majority voting. For equal-weight soft voting, direction-corrected scores are normalized using calibration-set statistics, converted into pseudo-probabilities, and then averaged. For calibration-weighted variants, base configurations are assigned deterministic non-negative weights computed only from the calibration split. We evaluate AUROC-based, AUPRC-based, TPR@1\%FPR-based, and rank-based weighting rules. If all weights become zero under a given rule, equal weights are used as a fallback.

The held-out test split is not used for selecting ensemble weights, score directions, normalization statistics, or thresholds. Thus, the voting variants remain training-free in the sense that no classifier, proxy language model, or supervised meta-classifier is trained.

\subsection{Thresholding and Evaluation Protocol}

Each baseline, individual wavelet-based method, or voting-ensemble variant produces either a continuous scalar score or an ensemble vote score for an input text. Since different scores may have different directions, we first determine the score direction on the calibration split. For each configuration, we compare the mean score of human-written texts and the mean score of LLM-generated texts in the calibration data. If the mean score of LLM-generated texts is greater than or equal to the mean score of human-written texts, the original score direction is preserved. Otherwise, the score is multiplied by $-1$ so that larger adjusted scores consistently indicate stronger evidence of LLM-generated text.

Let $s(X)$ denote the adjusted scalar score of an input text $X$. The final binary prediction is obtained by applying a threshold $\tau$:

\begin{equation}
\hat{y} =
\begin{cases}
1, & s(X) \geq \tau, \\
0, & s(X) < \tau,
\end{cases}
\end{equation}

where $\hat{y}=1$ denotes LLM-generated text and $\hat{y}=0$ denotes human-written text.

We use a stratified calibration-test evaluation protocol. For each dataset and class label, texts are randomly split into a calibration subset and a held-out test subset using a fixed random seed of 42. The calibration ratio is set to $0.30$, and the remaining $0.70$ of the data is reserved for final testing. This split is performed after duplicate removal and dataset normalization, and the same split assignment is used consistently across all evaluated score configurations.

The held-out test split is not used during score-direction alignment, threshold selection, or any configuration-specific decision-boundary adjustment. It is used only for final performance reporting after all calibration-based decisions have been fixed. This separation is important because it prevents the reported test performance from being inflated by selecting thresholds directly on the evaluation data.

The calibration split is used only for score-direction alignment, score normalization, deterministic ensemble weight computation, and threshold selection. It is not used to train a classifier, fine-tune a proxy language model, or learn a task-specific representation. This distinction is important because the proposed framework remains training-free: the calibration data only determine how a scalar score is oriented and where the decision threshold is placed.

The threshold $\tau$ is selected only on the calibration split. Specifically, we compute the precision-recall curve on the calibration scores and select the threshold that maximizes the F1 score:

\begin{equation}
F1 = \frac{2 \cdot P \cdot R}{P + R + \epsilon},
\end{equation}

where $P$ denotes precision, $R$ denotes recall, and $\epsilon$ is a small numerical constant used for stability. The selected threshold is then applied without modification to the held-out test split.

In this paper, we report the dataset-specific threshold setting. For each dataset, a separate threshold is selected using only its own calibration subset and then applied to the corresponding held-out test subset. This setting avoids mixing calibration distributions across datasets and provides dataset-level performance estimates. All results reported in the following sections are computed on the held-out 70\% test split after threshold selection on the 30\% calibration split.

A single threshold-dependent metric is not sufficient to characterize detector behavior, because performance can change depending on the selected decision threshold. Therefore, we report both threshold-independent and threshold-dependent metrics. Threshold-independent metrics evaluate the ranking quality and score separation before fixing a decision boundary, while threshold-dependent metrics evaluate the actual binary predictions after applying the calibration-selected threshold.

The threshold-independent metrics reported in this study are AUROC and AUPRC. AUROC and AUPRC are computed directly from the continuous adjusted scores and do not depend on the selected threshold.

Threshold-dependent metrics include accuracy and F1 score, which are computed after applying the calibration-selected threshold to the held-out test scores.

To evaluate performance under strict false-positive constraints, we also report true positive rate at fixed false positive rates. For a target false positive rate $\alpha$, we compute the ROC curve and select the largest valid point satisfying $\mathrm{FPR} \leq \alpha$. In particular, we report TPR@0.1\%FPR, TPR@1\%FPR, and TPR@5\%FPR. These metrics are useful for assessing whether a detector remains effective in operational settings where human-written texts must not be incorrectly flagged as AI-generated.

\section{Results}

This section presents the experimental results of the proposed wavelet-based and voting-ensemble framework for training-free LLM-generated text detection. We first compare individual wavelet-domain scores and voting-ensemble variants with statistical zero-shot baselines across HC3, M4, and MAGE. We then analyze performance under low-false-positive operating points, compare DWT-based scores with a DFT-based spectral energy baseline, and examine the effects of proxy language models and wavelet families.

\subsection{Overall Detection Performance}

This subsection reports the main detection results on the held-out test split. To make the comparison transparent, we report threshold-independent and threshold-dependent metrics in separate tables. Table~\ref{tab:threshold_independent_comparison} reports AUROC and AUPRC, while Table~\ref{tab:threshold_dependent_comparison} reports accuracy and F1 score.

For individual wavelet-domain scores, the proxy language model and wavelet family are selected according to AUROC on the held-out test split. For statistical baselines, the best proxy language model is also selected according to AUROC. For voting methods, we report the best-performing voting configuration for each dataset. The compared methods include mean log-likelihood, mean rank, mean log-rank, mean entropy, LRR, the three individual wavelet-domain scores, and four voting-ensemble variants: equal-weight hard voting, equal-weight soft voting, calibration-weighted hard voting, and calibration-weighted soft voting.

\begin{table}[!t]
\centering
\caption{Threshold-independent comparison among statistical baselines, individual wavelet-domain scores, and voting-ensemble variants on the held-out test split. Bold values indicate the best result within each dataset and metric.}
\label{tab:threshold_independent_comparison}
\scriptsize
\resizebox{\linewidth}{!}{%
\begin{tabular}{lllrr}
\toprule
Dataset & Method Group & Method & AUROC & AUPRC \\
\midrule
HC3 & Ensemble & Calibration-weighted hard voting & \textbf{0.9919} & \textbf{0.9914} \\
HC3 & Ensemble & Equal-weight hard voting & 0.9894 & 0.9912 \\
HC3 & Ensemble & Calibration-weighted soft voting & 0.9892 & 0.9882 \\
HC3 & Ensemble & Equal-weight soft voting & 0.9890 & 0.9910 \\
HC3 & Baseline & mean\_logrank, GPT-Neo & 0.9876 & 0.9868 \\
HC3 & Wavelet & multilevel\_energy, GPT-Neo, db1 & 0.9872 & 0.9871 \\
HC3 & Baseline & mean\_log\_likelihood, GPT-Neo & 0.9861 & 0.9849 \\
HC3 & Baseline & lrr\_score, GPT-J & 0.9861 & 0.9839 \\
HC3 & Wavelet & energy\_norm, GPT-J, coif1 & 0.9825 & 0.9752 \\
HC3 & Baseline & mean\_entropy, LLaMA-3 & 0.9562 & 0.9083 \\
HC3 & Baseline & mean\_rank, GPT-Neo & 0.9274 & 0.9342 \\
HC3 & Wavelet & window\_std, Falcon, db2 & 0.7137 & 0.5835 \\
\midrule
M4 & Ensemble & Calibration-weighted hard voting & \textbf{0.8477} & 0.8603 \\
M4 & Ensemble & Equal-weight hard voting & 0.8461 & 0.8477 \\
M4 & Ensemble & Calibration-weighted soft voting & 0.8420 & \textbf{0.8626} \\
M4 & Ensemble & Equal-weight soft voting & 0.8190 & 0.8506 \\
M4 & Wavelet & multilevel\_energy, LLaMA-3, db2 & 0.8185 & 0.8496 \\
M4 & Wavelet & energy\_norm, LLaMA-3, db4 & 0.7993 & 0.8257 \\
M4 & Baseline & mean\_logrank, Falcon & 0.7751 & 0.7363 \\
M4 & Baseline & lrr\_score, LLaMA-3 & 0.7743 & 0.8213 \\
M4 & Baseline & mean\_log\_likelihood, Falcon & 0.7680 & 0.7238 \\
M4 & Wavelet & window\_std, LLaMA-3, db2 & 0.7324 & 0.7696 \\
M4 & Baseline & mean\_rank, GPT-Neo & 0.6742 & 0.6319 \\
M4 & Baseline & mean\_entropy, LLaMA-3 & 0.6695 & 0.6680 \\
\midrule
MAGE & Ensemble & Calibration-weighted hard voting & \textbf{0.7471} & \textbf{0.5556} \\
MAGE & Ensemble & Equal-weight hard voting & 0.7314 & 0.5386 \\
MAGE & Ensemble & Calibration-weighted soft voting & 0.7165 & 0.5026 \\
MAGE & Ensemble & Equal-weight soft voting & 0.7139 & 0.4780 \\
MAGE & Wavelet & energy\_norm, GPT-Neo, db4 & 0.7138 & 0.4770 \\
MAGE & Wavelet & multilevel\_energy, GPT-Neo, db4 & 0.7068 & 0.4983 \\
MAGE & Baseline & lrr\_score, GPT-Neo & 0.6907 & 0.4739 \\
MAGE & Baseline & mean\_rank, GPT-Neo & 0.6402 & 0.4234 \\
MAGE & Baseline & mean\_logrank, GPT-Neo & 0.6378 & 0.4257 \\
MAGE & Wavelet & window\_std, GPT-Neo, coif1 & 0.6201 & 0.4638 \\
MAGE & Baseline & mean\_log\_likelihood, GPT-Neo & 0.6179 & 0.4070 \\
MAGE & Baseline & mean\_entropy, GPT-Neo & 0.6008 & 0.4144 \\
\bottomrule
\end{tabular}%
}
\end{table}

Table~\ref{tab:threshold_independent_comparison} shows that the voting-ensemble variants improve threshold-independent performance across all three datasets. On HC3, calibration-weighted hard voting achieves the highest AUROC and AUPRC, reaching 0.9919 and 0.9914, respectively. Although the strongest individual wavelet score remains highly competitive with the best statistical baseline, the ensemble variants provide a consistent additional improvement.

On M4, the benefit of voting-based fusion is more pronounced. The best individual wavelet configuration reaches an AUROC of 0.8185 and an AUPRC of 0.8496, already outperforming the evaluated statistical baselines in AUROC. Calibration-weighted hard voting further improves AUROC to 0.8477, while calibration-weighted soft voting gives the highest AUPRC of 0.8626. This indicates that combining multiple wavelet configurations can strengthen the ranking quality of the detector under heterogeneous benchmark conditions.

On MAGE, which is the most challenging dataset in our evaluation, calibration-weighted hard voting provides the strongest threshold-independent performance, reaching an AUROC of 0.7471 and an AUPRC of 0.5556. This represents a clear improvement over the best individual wavelet score and over the strongest statistical baseline. Overall, the results show that individual wavelet-domain scores provide competitive training-free detection signals, while calibration-guided voting further improves AUROC and AUPRC without training a supervised meta-classifier.

\begin{table}[!t]
\centering
\caption{Threshold-dependent comparison among statistical baselines, individual wavelet-domain scores, and voting-ensemble variants on the held-out test split. Bold values indicate the best result within each dataset and metric.}
\label{tab:threshold_dependent_comparison}
\scriptsize
\resizebox{\linewidth}{!}{%
\begin{tabular}{lllrr}
\toprule
Dataset & Method Group & Method & Accuracy & F1 \\
\midrule
HC3 & Ensemble & Equal-weight soft voting & \textbf{0.9727} & \textbf{0.9723} \\
HC3 & Ensemble & Calibration-weighted soft voting & 0.9720 & 0.9718 \\
HC3 & Baseline & mean\_logrank, GPT-Neo & 0.9717 & 0.9713 \\
HC3 & Ensemble & Equal-weight hard voting & 0.9710 & 0.9703 \\
HC3 & Ensemble & Calibration-weighted hard voting & 0.9703 & 0.9700 \\
HC3 & Baseline & mean\_log\_likelihood, GPT-Neo & 0.9686 & 0.9682 \\
HC3 & Wavelet & multilevel\_energy, GPT-Neo, db1 & 0.9680 & 0.9675 \\
HC3 & Baseline & lrr\_score, GPT-J & 0.9649 & 0.9644 \\
HC3 & Wavelet & energy\_norm, GPT-J, coif1 & 0.9566 & 0.9567 \\
HC3 & Baseline & mean\_entropy, LLaMA-3 & 0.9327 & 0.9337 \\
HC3 & Baseline & mean\_rank, GPT-Neo & 0.8577 & 0.8539 \\
HC3 & Wavelet & window\_std, Falcon, db2 & 0.7160 & 0.7640 \\
\midrule
M4 & Ensemble & Equal-weight hard voting & 0.7603 & \textbf{0.7747} \\
M4 & Ensemble & Calibration-weighted soft voting & 0.7514 & 0.7623 \\
M4 & Ensemble & Calibration-weighted hard voting & \textbf{0.7610} & 0.7571 \\
M4 & Baseline & mean\_logrank, Falcon & 0.7255 & 0.7548 \\
M4 & Wavelet & multilevel\_energy, LLaMA-3, db2 & 0.7597 & 0.7506 \\
M4 & Ensemble & Equal-weight soft voting & 0.7573 & 0.7500 \\
M4 & Baseline & mean\_log\_likelihood, Falcon & 0.7178 & 0.7492 \\
M4 & Wavelet & energy\_norm, LLaMA-3, db4 & 0.7286 & 0.7299 \\
M4 & Baseline & lrr\_score, LLaMA-3 & 0.7141 & 0.7107 \\
M4 & Baseline & mean\_rank, GPT-Neo & 0.6275 & 0.6861 \\
M4 & Wavelet & window\_std, LLaMA-3, db2 & 0.6431 & 0.6789 \\
M4 & Baseline & mean\_entropy, LLaMA-3 & 0.5696 & 0.6754 \\
\midrule
MAGE & Ensemble & Equal-weight soft voting & 0.6246 & \textbf{0.6117} \\
MAGE & Ensemble & Calibration-weighted hard voting & 0.6075 & 0.6111 \\
MAGE & Wavelet & energy\_norm, GPT-Neo, db4 & 0.6240 & 0.6072 \\
MAGE & Ensemble & Calibration-weighted soft voting & 0.6334 & 0.6053 \\
MAGE & Baseline & lrr\_score, GPT-Neo & 0.5815 & 0.5977 \\
MAGE & Wavelet & multilevel\_energy, GPT-Neo, db4 & \textbf{0.6386} & 0.5955 \\
MAGE & Ensemble & Equal-weight hard voting & 0.5459 & 0.5925 \\
MAGE & Baseline & mean\_logrank, GPT-Neo & 0.5433 & 0.5917 \\
MAGE & Baseline & mean\_log\_likelihood, GPT-Neo & 0.5373 & 0.5887 \\
MAGE & Baseline & mean\_rank, GPT-Neo & 0.5295 & 0.5674 \\
MAGE & Baseline & mean\_entropy, GPT-Neo & 0.4568 & 0.5407 \\
MAGE & Wavelet & window\_std, GPT-Neo, coif1 & 0.5811 & 0.5192 \\
\bottomrule
\end{tabular}%
}
\end{table}

Table~\ref{tab:threshold_dependent_comparison} reports the performance obtained after applying calibration-selected thresholds. On HC3, equal-weight soft voting achieves the highest accuracy and F1 score, slightly outperforming the strongest statistical baseline and the best individual wavelet score. This shows that ensemble fusion can also improve threshold-dependent performance on the easier HC3 benchmark.

On M4, equal-weight hard voting achieves the highest F1 score, while calibration-weighted hard voting gives the highest accuracy. The best individual wavelet score remains competitive, but the voting variants provide the strongest threshold-dependent results overall. On MAGE, equal-weight soft voting gives the highest F1 score, whereas the multilevel wavelet score achieves the highest accuracy. These results indicate that the best ensemble choice may depend on the target metric: calibration-weighted voting is strongest for AUROC and AUPRC, while equal-weight voting can sometimes produce stronger threshold-dependent F1 scores.

\subsection{Low-FPR Operating Point Analysis}

We further evaluate the detectors under strict false-positive-rate constraints. This analysis is important because practical LLM-generated text detection systems should minimize the risk of incorrectly flagging human-written text as AI-generated. Table~\ref{tab:low_fpr_comparison} reports TPR@0.1\%FPR, TPR@1\%FPR, and TPR@5\%FPR for the statistical baselines, individual wavelet-domain scores, and voting-ensemble variants evaluated on the held-out test split.

\begin{table}[!t]
\centering
\caption{Low-FPR operating point comparison among statistical baselines, individual wavelet-domain scores, and voting-ensemble variants on the held-out test split. Bold values indicate the best result within each dataset and metric.}
\label{tab:low_fpr_comparison}
\scriptsize
\resizebox{\linewidth}{!}{%
\begin{tabular}{lllrrr}
\toprule
Dataset & Method Group & Method & TPR@0.1\%FPR & TPR@1\%FPR & TPR@5\%FPR \\
\midrule
HC3 & Ensemble & Calibration-weighted hard voting & 0.4842 & \textbf{0.9425} & 0.9790 \\
HC3 & Ensemble & Equal-weight hard voting & \textbf{0.4946} & 0.9416 & 0.9793 \\
HC3 & Ensemble & Equal-weight soft voting & 0.4843 & 0.9374 & 0.9807 \\
HC3 & Ensemble & Calibration-weighted soft voting & 0.1884 & 0.9321 & 0.9811 \\
HC3 & Wavelet & multilevel\_energy, GPT-Neo, db1 & 0.1979 & 0.9067 & 0.9782 \\
HC3 & Baseline & mean\_logrank, GPT-Neo & 0.0808 & 0.9018 & \textbf{0.9817} \\
HC3 & Baseline & lrr\_score, GPT-J & 0.1207 & 0.9003 & 0.9713 \\
HC3 & Baseline & mean\_log\_likelihood, GPT-Neo & 0.1067 & 0.8413 & 0.9799 \\
HC3 & Wavelet & energy\_norm, GPT-J, coif1 & 0.0045 & 0.7539 & 0.9634 \\
HC3 & Baseline & mean\_rank, GPT-Neo & 0.0593 & 0.5769 & 0.7594 \\
HC3 & Baseline & mean\_entropy, LLaMA-3 & 0.0001 & 0.0172 & 0.8489 \\
HC3 & Wavelet & window\_std, Falcon, db2 & 0.0000 & 0.0000 & 0.0000 \\
\midrule
M4 & Ensemble & Calibration-weighted soft voting & \textbf{0.1275} & \textbf{0.3254} & 0.5317 \\
M4 & Baseline & lrr\_score, LLaMA-3 & 0.1151 & 0.3192 & 0.4787 \\
M4 & Ensemble & Equal-weight soft voting & 0.0637 & 0.3173 & 0.5418 \\
M4 & Wavelet & multilevel\_energy, LLaMA-3, db2 & 0.0679 & 0.3139 & 0.5379 \\
M4 & Ensemble & Calibration-weighted hard voting & 0.1238 & 0.2618 & 0.5262 \\
M4 & Wavelet & energy\_norm, LLaMA-3, db4 & 0.0440 & 0.2374 & 0.4640 \\
M4 & Wavelet & window\_std, LLaMA-3, db2 & 0.0649 & 0.2036 & 0.3558 \\
M4 & Ensemble & Equal-weight hard voting & 0.0636 & 0.1561 & \textbf{0.5428} \\
M4 & Baseline & mean\_entropy, LLaMA-3 & 0.0038 & 0.0626 & 0.2001 \\
M4 & Baseline & mean\_logrank, Falcon & 0.0106 & 0.0462 & 0.2122 \\
M4 & Baseline & mean\_log\_likelihood, Falcon & 0.0089 & 0.0396 & 0.1871 \\
M4 & Baseline & mean\_rank, GPT-Neo & 0.0020 & 0.0232 & 0.1328 \\
\midrule
MAGE & Baseline & mean\_entropy, GPT-Neo & \textbf{0.0053} & \textbf{0.0230} & 0.0684 \\
MAGE & Baseline & lrr\_score, GPT-Neo & 0.0050 & 0.0222 & 0.0780 \\
MAGE & Baseline & mean\_logrank, GPT-Neo & 0.0045 & 0.0200 & 0.0650 \\
MAGE & Wavelet & window\_std, GPT-Neo, coif1 & 0.0009 & 0.0178 & 0.1157 \\
MAGE & Baseline & mean\_log\_likelihood, GPT-Neo & 0.0042 & 0.0177 & 0.0571 \\
MAGE & Ensemble & Equal-weight hard voting & 0.0003 & 0.0119 & 0.1523 \\
MAGE & Baseline & mean\_rank, GPT-Neo & 0.0026 & 0.0114 & 0.0495 \\
MAGE & Ensemble & Calibration-weighted hard voting & 0.0003 & 0.0113 & \textbf{0.1775} \\
MAGE & Ensemble & Calibration-weighted soft voting & 0.0002 & 0.0090 & 0.0921 \\
MAGE & Wavelet & multilevel\_energy, GPT-Neo, db4 & 0.0005 & 0.0088 & 0.0914 \\
MAGE & Ensemble & Equal-weight soft voting & 0.0006 & 0.0071 & 0.0534 \\
MAGE & Wavelet & energy\_norm, GPT-Neo, db4 & 0.0005 & 0.0061 & 0.0533 \\
\bottomrule
\end{tabular}%
}
\end{table}

Table~\ref{tab:low_fpr_comparison} shows that low-FPR behavior varies substantially across datasets and does not always follow the same ranking as AUROC or AUPRC. On HC3, voting-based ensembles provide the strongest performance under strict false-positive constraints. Calibration-weighted hard voting achieves the highest TPR@1\%FPR, reaching 0.9425, while equal-weight hard voting gives the highest TPR@0.1\%FPR. These results indicate that ensemble fusion improves not only threshold-independent ranking performance but also high-confidence detection on the relatively easier HC3 benchmark.

On M4, calibration-weighted soft voting gives the best low-FPR performance, reaching 0.3254 at TPR@1\%FPR and 0.1275 at TPR@0.1\%FPR. This slightly improves over the LRR baseline at TPR@1\%FPR and also remains competitive with the best individual wavelet score. However, calibration-weighted hard voting, which gives the highest AUROC in Table~\ref{tab:threshold_independent_comparison}, does not produce the best TPR@1\%FPR on this dataset. This shows that the ensemble variant with the strongest overall ranking performance is not necessarily the best choice under strict false-positive constraints.

On MAGE, all methods show limited performance under strict low-FPR constraints. The mean entropy and LRR baselines obtain the highest TPR values at 0.1\%FPR and 1\%FPR, whereas calibration-weighted hard voting gives the highest TPR@5\%FPR. This indicates that the proposed ensemble improves threshold-independent separation on MAGE, as shown by AUROC and AUPRC, but reliable detection at extremely low false-positive rates remains challenging. Overall, the low-FPR results suggest that voting ensembles are beneficial on HC3 and M4, while MAGE remains a difficult out-of-distribution benchmark for training-free detection under strict operational constraints.

\subsection{Comparison with DFT-Based Spectral Energy}

Since the proposed framework analyzes token log-probability sequences in the wavelet domain, we further compare the individual DWT-based wavelet scores with a DFT-based spectral energy baseline. This comparison focuses specifically on signal-based zero-shot detection methods. Both DWT and DFT operate on the token log-probability sequence, but they represent the signal differently. The DFT baseline summarizes global frequency-domain energy, whereas the DWT-based wavelet scores capture localized and multiresolution variations through wavelet detail coefficients.

Table~\ref{tab:dwt_vs_dft} reports the three individual wavelet-domain scores and the DFT total energy baseline for each dataset on the held-out test split. For each wavelet-domain score, the best proxy language model and wavelet family are selected according to AUROC. For the DFT baseline, the best proxy language model is also selected according to AUROC. In addition to AUROC, AUPRC, and F1, we include TPR@1\%FPR as a representative low-false-positive operating point in this signal-based comparison.

\begin{table}[!b]
\centering
\caption{Signal-based comparison between the three individual DWT-based wavelet scores and the DFT total energy baseline on the held-out test split. Bold values indicate the best result within each dataset and metric.}
\label{tab:dwt_vs_dft}
\scriptsize
\resizebox{\linewidth}{!}{%
\begin{tabular}{lllccrrrr}
\toprule
Dataset & Method & Score & Proxy LM & Wavelet & AUROC & AUPRC & F1 & TPR@1\%FPR \\
\midrule
HC3 & DWT & energy\_norm & GPT-J & coif1 & 0.9825 & 0.9752 & 0.9567 & 0.7539 \\
HC3 & DWT & multilevel\_energy & GPT-Neo & db1 & \textbf{0.9872} & \textbf{0.9871} & \textbf{0.9675} & \textbf{0.9067} \\
HC3 & DWT & window\_std & Falcon & db2 & 0.7137 & 0.5835 & 0.7640 & 0.0000 \\
HC3 & DFT & dft\_total\_energy & LLaMA-3 & -- & 0.8131 & 0.6806 & 0.8305 & 0.0025 \\
\midrule
M4 & DWT & energy\_norm & LLaMA-3 & db4 & 0.7993 & 0.8257 & 0.7299 & 0.2374 \\
M4 & DWT & multilevel\_energy & LLaMA-3 & db2 & \textbf{0.8185} & \textbf{0.8496} & \textbf{0.7506} & \textbf{0.3139} \\
M4 & DWT & window\_std & LLaMA-3 & db2 & 0.7324 & 0.7696 & 0.6789 & 0.2036 \\
M4 & DFT & dft\_total\_energy & Falcon & -- & 0.8045 & 0.8048 & 0.7391 & 0.1411 \\
\midrule
MAGE & DWT & energy\_norm & GPT-Neo & db4 & \textbf{0.7138} & 0.4770 & \textbf{0.6072} & 0.0061 \\
MAGE & DWT & multilevel\_energy & GPT-Neo & db4 & 0.7068 & \textbf{0.4983} & 0.5955 & 0.0088 \\
MAGE & DWT & window\_std & GPT-Neo & coif1 & 0.6201 & 0.4638 & 0.5192 & \textbf{0.0178} \\
MAGE & DFT & dft\_total\_energy & GPT-Neo & -- & 0.5831 & 0.4191 & 0.5174 & 0.0101 \\
\bottomrule
\end{tabular}%
}
\end{table}

The results show that the strongest individual DWT-based score outperforms the DFT total energy baseline in AUROC across all three datasets. On HC3, multilevel energy reaches an AUROC of 0.9872, compared with 0.8131 for DFT total energy, giving an improvement of 0.1741. On M4, multilevel energy achieves an AUROC of 0.8185, while DFT total energy reaches 0.8045, corresponding to a smaller but positive improvement of 0.0140. On MAGE, first-level normalized detail energy reaches an AUROC of 0.7138, compared with 0.5831 for DFT total energy, giving an improvement of 0.1307.

The strongest individual DWT-based score also outperforms the DFT baseline in AUPRC and F1 across all three datasets. In terms of the representative low-FPR operating point, the strongest DWT-based score gives higher TPR@1\%FPR than the DFT baseline on HC3 and M4. On MAGE, window-energy variability gives the highest TPR@1\%FPR among the signal-based scores, although the absolute value remains low. Overall, these results suggest that localized multiresolution analysis provides a stronger signal-based representation than global spectral energy alone. The difference is especially large on HC3 and MAGE, indicating that DFT total energy may miss discriminative local fluctuations in the token log-probability sequence. The smaller gap on M4 shows that DFT total energy is a competitive signal-domain baseline on this dataset, but the DWT-based multilevel energy score still provides the strongest individual signal-based performance.

\subsection{Proxy Model and Wavelet Family Sensitivity}

\begin{figure}[!b]
    \centering
    \includegraphics[width=\linewidth]{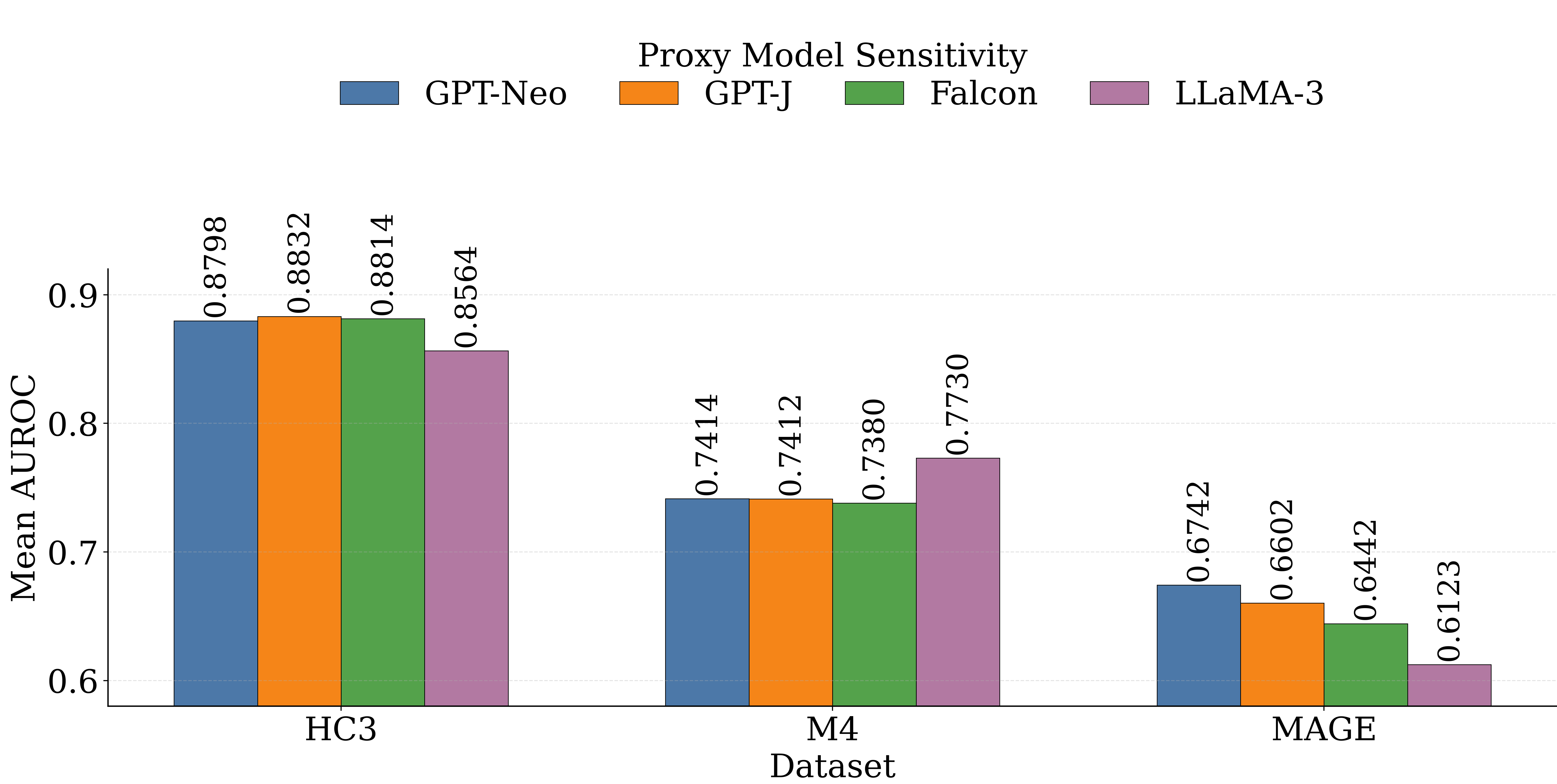}
    \caption{Proxy model sensitivity of the proposed wavelet-based detector on the held-out test split.}
    \label{fig:proxy_model_sensitivity}
\end{figure}

We next analyze how proxy language models and wavelet families affect detection performance on the held-out test split. Instead of reporting only the single best configuration, we use mean AUROC values to obtain a more stable sensitivity analysis. For proxy model sensitivity, each bar in Figure~\ref{fig:proxy_model_sensitivity} represents the mean AUROC over the three scalar wavelet scores and five wavelet families for a fixed dataset and proxy model.

\begin{figure}[!t]
    \centering
    \includegraphics[width=\linewidth]{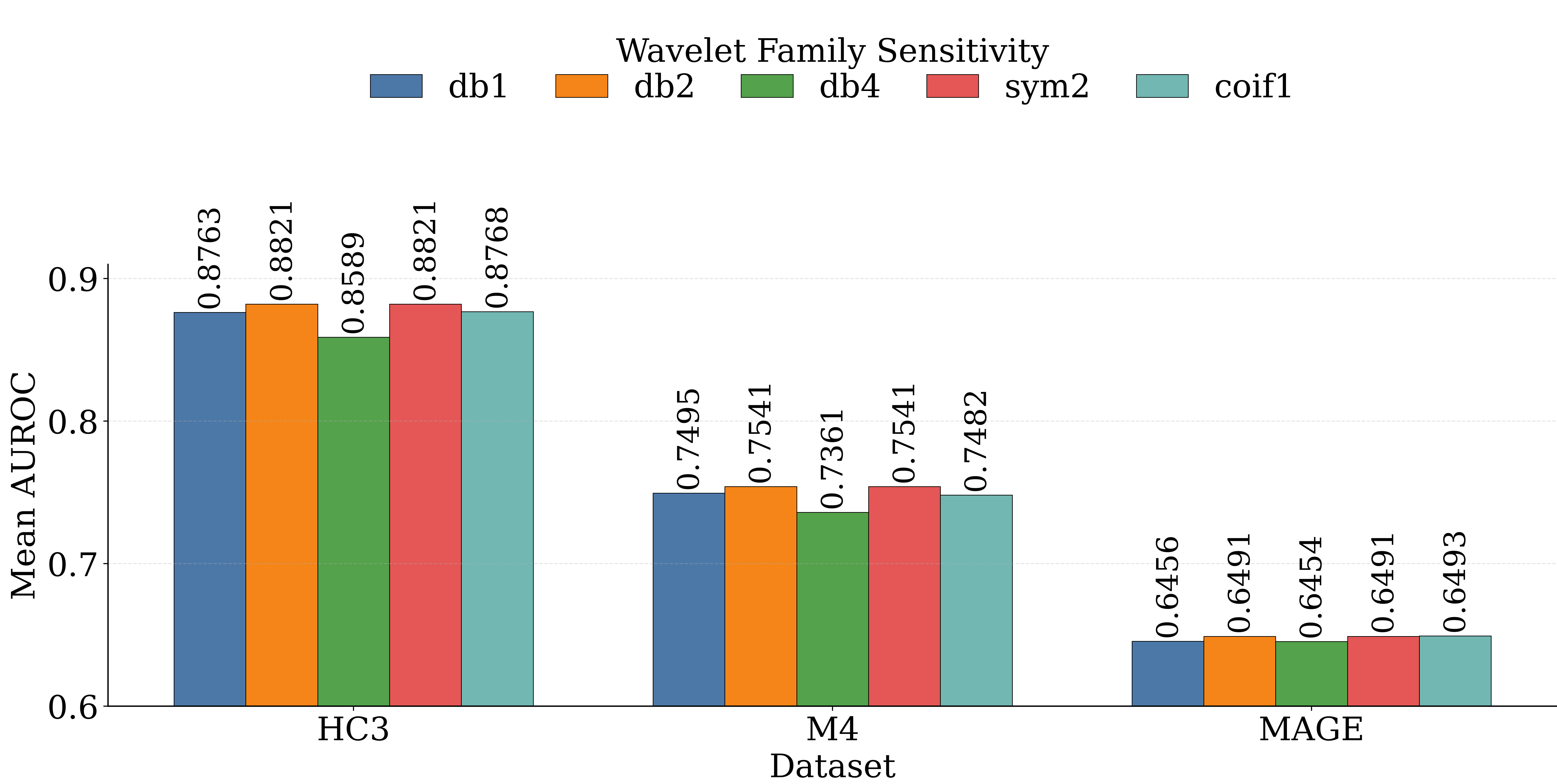}
    \caption{Wavelet family sensitivity of the proposed method on the held-out test split.}
    \label{fig:wavelet_family_sensitivity}
\end{figure}

Figure~\ref{fig:proxy_model_sensitivity} shows that proxy model choice affects the proposed method differently across datasets. On HC3, the proxy models produce relatively close mean AUROC values, with GPT-J-6B obtaining the highest mean AUROC of 0.8832. On M4, LLaMA-3-8B is clearly the strongest proxy model, reaching a mean AUROC of 0.7730. On MAGE, GPT-Neo-2.7B performs best with a mean AUROC of 0.6742. These results indicate that proxy model choice becomes more important on heterogeneous datasets such as M4 and MAGE.

For wavelet family sensitivity, each bar in Figure~\ref{fig:wavelet_family_sensitivity} represents the mean AUROC over the four proxy language models and three scalar wavelet scores for a fixed dataset and wavelet family. This analysis shows whether a wavelet family performs consistently across different proxy models and score definitions.

Figure~\ref{fig:wavelet_family_sensitivity} shows that the average effect of the wavelet family is moderate. On HC3, db2 and sym2 obtain the highest mean AUROC values, both reaching 0.8821. On M4, db2 and sym2 again perform best, both reaching 0.7541. On MAGE, the wavelet families are closely grouped, with coif1 obtaining the highest mean AUROC of 0.6493. Overall, proxy model choice appears to have a stronger effect than wavelet family choice, especially on M4 and MAGE.

% -------------------------------------------------
% 5. Discussion
% -------------------------------------------------
\section{Discussion}

This section discusses the main findings of the experimental evaluation in terms of threshold-independent performance, threshold-dependent performance, low-false-positive operating points, signal-based comparison with DFT, sensitivity to proxy language models and wavelet families, and the limitations of the proposed framework.

\noindent\textbf{Main Findings.}
The results show that individual wavelet-domain scores provide competitive training-free detection signals, while voting-based ensemble variants further improve the strongest threshold-independent performance across all three datasets. On HC3, the best individual wavelet configuration reaches an AUROC of 0.9872, which is very close to the strongest statistical baseline, mean log-rank, with an AUROC of 0.9876. However, calibration-weighted hard voting further improves AUROC to 0.9919 and AUPRC to 0.9914. This indicates that although HC3 is already well separated by global token-level statistics, combining multiple wavelet configurations can still provide a small but consistent improvement.

For the M4 dataset, the advantage of the proposed framework is more pronounced. The best individual wavelet score, multilevel detail energy, reaches an AUROC of 0.8185 and an AUPRC of 0.8496, outperforming the evaluated statistical baselines in threshold-independent performance. When voting-based fusion is applied, calibration-weighted hard voting further improves AUROC to 0.8477, while calibration-weighted soft voting obtains the highest AUPRC of 0.8626. These results suggest that heterogeneous benchmark conditions benefit from combining complementary wavelet configurations, because different proxy models, wavelet families, and score definitions may capture different aspects of token-level probability dynamics.

For the MAGE dataset, which is the most challenging benchmark in the evaluation, the best individual wavelet score reaches an AUROC of 0.7138, outperforming the strongest statistical baseline, LRR, which reaches 0.6907. Calibration-weighted hard voting further improves AUROC to 0.7471 and AUPRC to 0.5556. This improvement is important because MAGE includes more heterogeneous generator and domain conditions, where individual scalar scores may be less stable. However, the absolute performance on MAGE remains lower than on HC3 and M4, showing that highly heterogeneous out-of-distribution detection remains challenging even when calibration-guided voting is used.

\noindent\textbf{Behavior of the wavelet-domain scores.}
The behavior of the three wavelet-domain scores provides useful insight into the type of signal captured by the proposed framework. Multilevel detail energy gives the best AUROC on HC3 and M4, suggesting that these datasets benefit from aggregating localized fluctuations across multiple decomposition levels. In contrast, first-level normalized detail energy gives the best AUROC on MAGE, which may indicate that short-scale normalized fluctuations provide a more stable signal under more heterogeneous and out-of-distribution conditions. Window-energy variability is generally weaker in terms of overall AUROC and threshold-dependent performance. However, it obtains the highest TPR@5\%FPR on MAGE among the wavelet scores, suggesting that local energy dispersion may still capture useful information at some operating points even when its overall ranking performance is limited.

\noindent\textbf{Why wavelet-domain scores help.}
The advantage of the proposed method comes from treating token log-probabilities as an ordered signal rather than as an unordered collection of scalar statistics. Global scores such as mean log-likelihood, mean rank, and mean entropy compress the entire text into a single aggregate value. Although such scores are effective in many cases, they discard information about where uncertainty, fluency shifts, or local probability fluctuations occur in the sequence. In contrast, wavelet decomposition preserves both scale and locality. Fine-scale detail coefficients capture abrupt token-level changes, while coarser detail coefficients summarize broader fluctuations across longer spans. Since LLM-generated and human-written texts may differ not only in average likelihood but also in the temporal structure of likelihood fluctuations, wavelet-domain scores can capture complementary information.

\noindent\textbf{Comparison with DFT-based spectral energy.}
The comparison with DFT total energy further supports this interpretation. Although both DFT and DWT treat the token log-probability sequence as a signal, the DFT baseline represents global frequency content and loses explicit localization. The strongest DWT-based score outperforms the DFT total energy baseline in AUROC on all three datasets, with improvements of 0.1741 on HC3, 0.0140 on M4, and 0.1307 on MAGE. The strongest DWT-based scores also outperform the DFT baseline in AUPRC and F1 across all three datasets. In the representative low-FPR operating point reported in the signal-based comparison, DWT-based scores also obtain higher TPR@1\%FPR values than the DFT baseline. These results suggest that where fluctuations occur in the sequence matters, not only how much global spectral energy the sequence contains.

\noindent\textbf{Sensitivity to proxy models and wavelet families.}
The results also show that the effectiveness of the proposed framework is affected by the choice of proxy language model. Unlike the main performance tables, the sensitivity analysis reports mean AUROC values across multiple wavelet scores and wavelet families, providing a more stable view of proxy-model behavior. On HC3, GPT-J-6B obtains the highest mean AUROC, although GPT-Neo-2.7B gives the best single wavelet configuration in the main comparison. On M4, LLaMA-3-8B is clearly the strongest proxy model in terms of mean AUROC. On MAGE, GPT-Neo-2.7B gives the highest mean AUROC. These results indicate that no single proxy model universally dominates across datasets. Instead, the proxy model affects the probability landscape from which the token log-probability signal is extracted, and different benchmarks may favor different proxy distributions.

A similar but more moderate pattern appears for wavelet families. In the sensitivity analysis, db2 and sym2 give the highest mean AUROC values on HC3 and M4, while coif1 gives the highest mean AUROC on MAGE. This differs from the best single configurations in the main comparison, where db1, db2, and db4 appear as the best wavelet families depending on the dataset and score. This distinction is important: the best single configuration identifies the strongest observed setting, whereas mean AUROC reflects average behavior across proxy models and score definitions. Overall, the results suggest that wavelet family choice matters, but proxy model choice has a stronger effect, especially on M4 and MAGE.

\noindent\textbf{Low-FPR behavior.}
The low-FPR results highlight an important distinction between overall ranking performance and operational reliability. On HC3, voting-based ensembles provide the strongest performance under strict false-positive constraints. Calibration-weighted hard voting achieves the highest TPR@1\%FPR, while equal-weight hard voting gives the highest TPR@0.1\%FPR. This indicates that ensemble fusion improves not only AUROC and AUPRC but also high-confidence detection on the relatively easier HC3 benchmark.

On M4, the low-FPR behavior is more nuanced. Calibration-weighted hard voting gives the highest AUROC, but calibration-weighted soft voting provides the best TPR@0.1\%FPR and TPR@1\%FPR. This shows that the ensemble variant with the strongest threshold-independent ranking performance is not necessarily the best choice under strict false-positive constraints. Soft voting may preserve more fine-grained score information than hard voting, which can be beneficial when the operating point is constrained by very low false-positive rates.

On MAGE, all methods remain limited under strict low-FPR constraints. Although calibration-weighted hard voting gives the highest AUROC, AUPRC, and TPR@5\%FPR, statistical baselines such as mean entropy and LRR achieve higher TPR values at 0.1\%FPR and 1\%FPR. Therefore, the MAGE results should not be interpreted as evidence of a reliable operational detector under extremely strict false-positive requirements. Instead, they show that wavelet-domain scores and voting ensembles improve overall ranking performance, while low-FPR detection on highly heterogeneous benchmarks remains an open challenge.

\noindent\textbf{Limitations and future work.}
Despite its training-free, interpretable, and signal-based design, the proposed framework has several limitations that should be considered when interpreting the results. The first limitation is the dependence on proxy language models. Since token log-probabilities are extracted from proxy models rather than from the true generators, detection performance can vary depending on how well the proxy model represents the probability structure of the evaluated texts. Although voting-based ensembles partly reduce this dependence by combining multiple proxy models and wavelet configurations, the results still show that proxy choice affects performance across datasets.

A second limitation concerns calibration. The proposed ensemble variants remain training-free because they do not train a classifier, fine-tune a language model, or learn a supervised meta-classifier. However, they still use a calibration split for score-direction correction, score normalization, threshold selection, and deterministic weight computation. This makes the method calibration-guided rather than completely calibration-free. Future work should investigate cross-dataset calibration, unsupervised calibration strategies, and more stable weighting rules that preserve low-FPR behavior under distribution shift.

Another limitation concerns input length and decomposition depth. In this study, all input texts are truncated to a maximum length of 512 tokens, and the DWT decomposition is limited to a maximum of three levels. These choices make the evaluation computationally feasible and consistent across proxy language models, but they may discard useful evidence from later portions of long texts or from deeper multiresolution structures. This issue could be addressed through sliding-window scoring, longer-context proxy models, or adaptive decomposition depths that depend on the length and structure of each input signal.

Low-FPR performance also remains limited on the MAGE dataset. This is important because real-world deployment often requires very low false-positive rates, especially in settings where incorrectly flagging human-written text may have serious consequences. Improving this operating region may require better score calibration, proxy-model selection, longer-context analysis, or hybrid approaches that combine wavelet-domain signals with complementary statistical indicators.

Finally, the evaluation is restricted to benchmark datasets. Although HC3, M4, and MAGE provide diverse and challenging test settings, real-world writing may involve editing, paraphrasing, mixed human-AI authorship, and domain-specific writing conventions. Evaluating wavelet-domain and voting-based detection under such realistic conditions remains an important direction for future research.

% -------------------------------------------------
% 6. Conclusion
% -------------------------------------------------
\section{Conclusion}

This paper presented a training-free wavelet-based and voting-ensemble framework for detecting LLM-generated text from token-level conditional log-probability signals. Instead of reducing a text only to global likelihood, rank, or entropy statistics, the proposed framework preserves the sequential structure of token probabilities and analyzes the resulting signal through discrete wavelet decomposition. This enables the detector to capture localized multiresolution variations that are not directly represented by conventional zero-shot scoring methods. In addition to individual wavelet-domain scores, we also evaluated equal-weight and calibration-weighted voting variants that combine multiple wavelet configurations without training a supervised meta-classifier.

The experimental results on HC3, M4, and MAGE show that individual wavelet-domain scores provide competitive and interpretable signals for LLM-generated text detection. The best single wavelet configurations achieve AUROC values of 0.9872 on HC3, 0.8185 on M4, and 0.7138 on MAGE. These results are competitive with strong statistical baselines on HC3 and outperform the evaluated statistical and DFT-based spectral baselines on M4 and MAGE in threshold-independent performance. The comparison with DFT total energy further shows that localized multiresolution analysis is more effective than global spectral energy across all three datasets.

The voting-based results further show that combining multiple wavelet configurations can improve detection performance, especially in terms of AUROC and AUPRC. Calibration-weighted hard voting achieves the best AUROC values of 0.9919 on HC3, 0.8477 on M4, and 0.7471 on MAGE. These improvements indicate that different proxy models, wavelet families, and wavelet-domain scores provide complementary detection signals. However, the low-FPR analysis also shows that the best AUROC configuration is not always the strongest option under strict false-positive constraints. In particular, MAGE remains challenging for reliable low-FPR detection.

Overall, the findings suggest that wavelet-domain analysis provides an effective, interpretable, and training-free signal for LLM-generated text detection, while calibration-guided voting can further improve threshold-independent performance without supervised detector training. Future work can extend this framework through improved calibration strategies, longer-context signal analysis, more stable low-FPR optimization, and hybrid detectors that combine wavelet-domain scores with complementary statistical signals. Evaluating the method under realistic writing conditions, such as paraphrased text, edited AI outputs, and mixed human-AI authorship, also remains an important direction for future research.

% -------------------------------------------------
% References
% -------------------------------------------------

\bibliographystyle{unsrt}
\bibliography{references}

@article{fraser2025detecting,
  title = {Detecting {AI}-Generated Text: Factors Influencing Detectability with Current Methods},
  author = {Fraser, Kathleen C. and Dawkins, Hillary and Kiritchenko, Svetlana},
  journal = {Journal of Artificial Intelligence Research},
  volume = {82},
  pages = {2233--2278},
  year = {2025},
  month = apr,
  publisher = {{AI} Access Foundation},
  issn = {1076-9757},
  doi = {10.1613/jair.1.16665},
  url = {http://dx.doi.org/10.1613/jair.1.16665}
}

@article{wu2024survey,
    title = "A Survey on {LLM}-Generated Text Detection: Necessity, Methods, and Future Directions",
    author = "Wu, Junchao  and
      Yang, Shu  and
      Zhan, Runzhe  and
      Yuan, Yulin  and
      Chao, Lidia Sam  and
      Wong, Derek Fai",
    journal = "Computational Linguistics",
    volume = "51",
    number = "1",
    month = mar,
    year = "2025",
    address = "Cambridge, MA",
    publisher = "MIT Press",
    url = "https://aclanthology.org/2025.cl-1.8/",
    doi = "10.1162/coli_a_00549",
    pages = "275--338"
}

@inproceedings{yang2023survey,
    title = "A Survey on Detection of {LLM}s-Generated Content",
    author = "Yang, Xianjun  and
      Pan, Liangming  and
      Zhao, Xuandong  and
      Chen, Haifeng  and
      Petzold, Linda Ruth  and
      Wang, William Yang  and
      Cheng, Wei",
    editor = "Al-Onaizan, Yaser  and
      Bansal, Mohit  and
      Chen, Yun-Nung",
    booktitle = "Findings of the Association for Computational Linguistics: EMNLP 2024",
    month = nov,
    year = "2024",
    address = "Miami, Florida, USA",
    publisher = "Association for Computational Linguistics",
    url = "https://aclanthology.org/2024.findings-emnlp.572/",
    doi = "10.18653/v1/2024.findings-emnlp.572",
    pages = "9786--9805"
}

@article{mallat1989wavelet,
  title = {A Theory for Multiresolution Signal Decomposition: The Wavelet Representation},
  author = {Mallat, S. G.},
  journal = {{IEEE} Transactions on Pattern Analysis and Machine Intelligence},
  volume = {11},
  number = {7},
  pages = {674--693},
  year = {1989},
  month = jul,
  publisher = {{IEEE} Computer Society},
  address = {USA},
  issn = {0162-8828},
  doi = {10.1109/34.192463},
  url = {https://doi.org/10.1109/34.192463}
}

@book{daubechies1992wavelets,
  title = {Ten Lectures on Wavelets},
  author = {Daubechies, Ingrid},
  publisher = {Society for Industrial and Applied Mathematics},
  year = {1992},
  doi = {10.1137/1.9781611970104},
  url = {https://epubs.siam.org/doi/abs/10.1137/1.9781611970104}
}

@INPROCEEDINGS{fu2025fdllm,
  author={Fu, Zhiyuan and Chen, Junfan and Zhang, Lan and Yang, Ting and Niu, Jun and Sun, Hongyu and Li, Ruidong and Liu, Peng and Wang, Jice and He, Fannv and Zhang, Yuqing},
  booktitle={2025 IEEE 24th International Conference on Trust, Security and Privacy in Computing and Communications (TrustCom)}, 
  title={FDLLM: A Dedicated Detector for Black-Box LLMs Fingerprinting}, 
  year={2025},
  volume={},
  number={},
  pages={1374-1379},
  doi={10.1109/Trustcom66490.2025.00159}}

@inproceedings{hao2025learning2rewrite,
    title = "Learning to Rewrite: Generalized {LLM}-Generated Text Detection",
    author = "Hao, Wei  and
      Li, Ran  and
      Zhao, Weiliang  and
      Yang, Junfeng  and
      Mao, Chengzhi",
    editor = "Che, Wanxiang  and
      Nabende, Joyce  and
      Shutova, Ekaterina  and
      Pilehvar, Mohammad Taher",
    booktitle = "Proceedings of the 63rd Annual Meeting of the Association for Computational Linguistics (Volume 1: Long Papers)",
    month = jul,
    year = "2025",
    address = "Vienna, Austria",
    publisher = "Association for Computational Linguistics",
    url = "https://aclanthology.org/2025.acl-long.322/",
    doi = "10.18653/v1/2025.acl-long.322",
    pages = "6421--6434",
    ISBN = "979-8-89176-251-0"
}

@inproceedings{hu2023radar,
author = {Hu, Xiaomengc and Chen, Pin-Yu and Ho, Tsung-Yi},
title = {RADAR: robust AI-text detection via adversarial learning},
year = {2023},
publisher = {Curran Associates Inc.},
address = {Red Hook, NY, USA},
booktitle = {Proceedings of the 37th International Conference on Neural Information Processing Systems},
articleno = {662},
numpages = {19},
location = {New Orleans, LA, USA},
series = {NIPS '23}
}

@article{zeng2025humanoutliers,
  title = {Human Texts Are Outliers: Detecting {LLM}-generated Texts via Out-of-distribution Detection},
  author = {Zeng, Cong and Tang, Shengkun and Chen, Yuanzhou and Shen, Zhiqiang and Yu, Wenchao and Zhao, Xujiang and Chen, Haifeng and Cheng, Wei and Xu, Zhiqiang},
  journal = {arXiv preprint arXiv:2510.08602},
  year = {2025},
  eprint = {2510.08602},
  archivePrefix = {arXiv},
  primaryClass = {cs.CL},
  doi = {10.48550/arXiv.2510.08602},
  url = {https://arxiv.org/abs/2510.08602}
}

@article{zheng2025lm2otifs,
  title = {{LM}$^2$otifs: An Explainable Framework for Machine-Generated Texts Detection},
  author = {Zheng, Xu and Chen, Zhuomin and Schafir, Esteban and Chen, Sipeng and Salehi, Hojat Allah and Chen, Haifeng and Shirani, Farhad and Cheng, Wei and Luo, Dongsheng},
  journal = {arXiv preprint arXiv:2505.12507},
  year = {2025},
  eprint = {2505.12507},
  archivePrefix = {arXiv},
  primaryClass = {cs.CL},
  doi = {10.48550/arXiv.2505.12507},
  url = {https://arxiv.org/abs/2505.12507}
}

@inproceedings{
bao2024fastdetectgpt,
title={Fast-Detect{GPT}: Efficient Zero-Shot Detection of Machine-Generated Text via Conditional Probability Curvature},
author={Guangsheng Bao and Yanbin Zhao and Zhiyang Teng and Linyi Yang and Yue Zhang},
booktitle={The Twelfth International Conference on Learning Representations},
year={2024},
url={https://openreview.net/forum?id=Bpcgcr8E8Z}
}

@inproceedings{hans2024binoculars,
author = {Hans, Abhimanyu and Schwarzschild, Avi and Cherepanova, Valeriia and Kazemi, Hamid and Saha, Aniruddha and Goldblum, Micah and Geiping, Jonas and Goldstein, Tom},
title = {Spotting LLMs with binoculars: zero-shot detection of machine-generated text},
year = {2024},
publisher = {JMLR.org},
booktitle = {Proceedings of the 41st International Conference on Machine Learning},
articleno = {698},
numpages = {19},
location = {Vienna, Austria},
series = {ICML'24}
}

@inproceedings{luo2026specdetect,
author = {Luo, Haitong and Zhang, Weiyao and Wang, Suhang and Zou, Wenji and Lin, Chungang and Meng, Xuying and Zhang, Yujun},
title = {SpecDetect: simple, fast, and training-free detection of LLM-generated text via spectral analysis},
year = {2026},
isbn = {978-1-57735-906-7},
publisher = {AAAI Press},
url = {https://doi.org/10.1609/aaai.v40i38.40510},
doi = {10.1609/aaai.v40i38.40510},
booktitle = {Proceedings of the Fortieth AAAI Conference on Artificial Intelligence and Thirty-Eighth Conference on Innovative Applications of Artificial Intelligence and Sixteenth Symposium on Educational Advances in Artificial Intelligence},
articleno = {3609},
numpages = {9},
series = {AAAI'26/IAAI'26/EAAI'26}
}

@inproceedings{mitchell2023detectgpt,
author = {Mitchell, Eric and Lee, Yoonho and Khazatsky, Alexander and Manning, Christopher D. and Finn, Chelsea},
title = {DetectGPT: zero-shot machine-generated text detection using probability curvature},
year = {2023},
publisher = {JMLR.org},
booktitle = {Proceedings of the 40th International Conference on Machine Learning},
articleno = {1038},
numpages = {13},
location = {Honolulu, Hawaii, USA},
series = {ICML'23}
}

@inproceedings{su2023detectllm,
    title = "{D}etect{LLM}: Leveraging Log Rank Information for Zero-Shot Detection of Machine-Generated Text",
    author = "Su, Jinyan  and
      Zhuo, Terry  and
      Wang, Di  and
      Nakov, Preslav",
    editor = "Bouamor, Houda  and
      Pino, Juan  and
      Bali, Kalika",
    booktitle = "Findings of the Association for Computational Linguistics: EMNLP 2023",
    month = dec,
    year = "2023",
    address = "Singapore",
    publisher = "Association for Computational Linguistics",
    url = "https://aclanthology.org/2023.findings-emnlp.827/",
    doi = "10.18653/v1/2023.findings-emnlp.827",
    pages = "12395--12412"
}

@article{sun2025textreorder,
  title = {Zero-shot Detection of {LLM}-Generated Text via Text Reorder},
  author = {Sun, Jingtao and Lv, Zhanglong},
  journal = {Neurocomputing},
  volume = {631},
  pages = {129829},
  year = {2025},
  issn = {0925-2312},
  doi = {10.1016/j.neucom.2025.129829},
  url = {https://www.sciencedirect.com/science/article/pii/S0925231225005016}
}

@inproceedings{wu2025gecscore,
    title = "Who Wrote This? The Key to Zero-Shot {LLM}-Generated Text Detection Is {GECS}core",
    author = "Wu, Junchao  and
      Zhan, Runzhe  and
      Wong, Derek F.  and
      Yang, Shu  and
      Liu, Xuebo  and
      Chao, Lidia S.  and
      Zhang, Min",
    editor = "Rambow, Owen  and
      Wanner, Leo  and
      Apidianaki, Marianna  and
      Al-Khalifa, Hend  and
      Eugenio, Barbara Di  and
      Schockaert, Steven",
    booktitle = "Proceedings of the 31st International Conference on Computational Linguistics",
    month = jan,
    year = "2025",
    address = "Abu Dhabi, UAE",
    publisher = "Association for Computational Linguistics",
    url = "https://aclanthology.org/2025.coling-main.684/",
    pages = "10275--10292"
}

@article{yang2025siltd,
  title = {{SILTD}: Structural Information for {LLM}-Generated Text Detection},
  author = {Yang, Jing and Wang, Shi and Zi, Kangli and Sun, Yanshun and Huang, Yuwei and Luo, Tianyu},
  journal = {International Journal of Machine Learning and Cybernetics},
  volume = {16},
  number = {9},
  pages = {6095--6110},
  year = {2025},
  issn = {1868-808X},
  doi = {10.1007/s13042-025-02616-x},
  url = {https://doi.org/10.1007/s13042-025-02616-x}
}

@inproceedings{chen2025divscore,
    title = "{D}iv{S}core: Zero-Shot Detection of {LLM}-Generated Text in Specialized Domains",
    author = "Chen, Zhihui  and
      He, Kai  and
      Huang, Yucheng  and
      Zhu, Yunxiao  and
      Feng, Mengling",
    editor = "Christodoulopoulos, Christos  and
      Chakraborty, Tanmoy  and
      Rose, Carolyn  and
      Peng, Violet",
    booktitle = "Proceedings of the 2025 Conference on Empirical Methods in Natural Language Processing",
    month = nov,
    year = "2025",
    address = "Suzhou, China",
    publisher = "Association for Computational Linguistics",
    url = "https://aclanthology.org/2025.emnlp-main.971/",
    doi = "10.18653/v1/2025.emnlp-main.971",
    pages = "19231--19253",
    ISBN = "979-8-89176-332-6"
}

@inproceedings{guo2024detective,
author = {Guo, Xun and Zhang, Shan and He, Yongxin and Zhang, Ting and Feng, Wanquan and Huang, Haibin and Ma, Chongyang},
title = {DeTeCtive: detecting AI-generated text via multi-level contrastive learning},
year = {2024},
isbn = {9798331314385},
publisher = {Curran Associates Inc.},
address = {Red Hook, NY, USA},
booktitle = {Proceedings of the 38th International Conference on Neural Information Processing Systems},
articleno = {2802},
numpages = {28},
location = {Vancouver, BC, Canada},
series = {NIPS '24}
}

@article{he2025detree,
  title = {{DETree}: {DE}tecting Human-{AI} Collaborative Texts via Tree-Structured Hierarchical Representation Learning},
  author = {He, Yongxin and Zhang, Shan and Cao, Yixuan and Ma, Lei and Luo, Ping},
  journal = {arXiv preprint arXiv:2510.17489},
  year = {2025},
  eprint = {2510.17489},
  archivePrefix = {arXiv},
  primaryClass = {cs.CL},
  doi = {10.48550/arXiv.2510.17489},
  url = {https://arxiv.org/abs/2510.17489}
}

@article{bao2025hart2d,
  title = {Decoupling Content and Expression: Two-Dimensional Detection of {AI}-Generated Text},
  author = {Bao, Guangsheng and Rong, Lihua and Zhao, Yanbin and Zhou, Qiji and Zhang, Yue},
  journal = {arXiv preprint arXiv:2503.00258},
  year = {2025},
  eprint = {2503.00258},
  archivePrefix = {arXiv},
  primaryClass = {cs.CL},
  doi = {10.48550/arXiv.2503.00258},
  url = {https://arxiv.org/abs/2503.00258}
}

@inproceedings{cheng2024biscope,
  title = {{BiScope}: {AI}-generated Text Detection by Checking Memorization of Preceding Tokens},
  author = {Cheng, Siyuan and Guo, Hanxi and Jin, Xiaolong and Shen, Guangyu and Tao, Guanhong and Zhang, Kaiyuan and Zhang, Xiangyu and Zhang, Zhuo},
  booktitle = {Advances in Neural Information Processing Systems},
  volume = {37},
  pages = {104065--104090},
  year = {2024},
  publisher = {Neural Information Processing Systems Foundation, Inc.},
  doi = {10.5220/79017-3307},
  url = {https://openreview.net/forum?id=Hew2JSDycr}
}

@article{le2025metricdet,
  title = {A Metric-Based Detection System for Large Language Model Texts},
  author = {Le, Linh and Tran, Dung},
  journal = {{ACM} Transactions on Management Information Systems},
  volume = {16},
  number = {1},
  articleno = {8},
  numpages = {19},
  year = {2025},
  month = feb,
  publisher = {Association for Computing Machinery},
  address = {New York, NY, USA},
  issn = {2158-656X},
  doi = {10.1145/3704739},
  url = {https://doi.org/10.1145/3704739}
}

@inproceedings{li2025continualorigin,
  title = {Continual Origin Tracing of {LLM}-Generated Text},
  author = {Li, Haoran and Wang, Quan},
  booktitle = {Proceedings of the 48th International {ACM} {SIGIR} Conference on Research and Development in Information Retrieval},
  pages = {479--489},
  year = {2025},
  series = {{SIGIR} '25},
  publisher = {Association for Computing Machinery},
  address = {New York, NY, USA},
  location = {Padua, Italy},
  isbn = {9798400715921},
  doi = {10.1145/3726302.3729935},
  url = {https://doi.org/10.1145/3726302.3729935}
}

@inproceedings{titze2025logaid,
  title = {{LOG-AID}: Logit-Based Statistical Features for {AI} Text Detection},
  author = {Titze, Sophie and Halvani, Oren},
  booktitle = {Notebook for {PAN} at {CLEF} 2025},
  year = {2025},
  location = {Madrid, Spain},
  doi = {10.24406/publica-5954},
  url = {https://publica.fraunhofer.de/handle/publica/497951}
}

@article{wang2025benatten,
  title = {Can Attention Detect {AI}-Generated Text? A Novel {Benford}'s Law-Based Approach},
  author = {Wang, Zhenhua and Xu, Guang and Ren, Ming},
  journal = {Information Processing \& Management},
  volume = {62},
  number = {4},
  pages = {104139},
  year = {2025},
  issn = {0306-4573},
  doi = {10.1016/j.ipm.2025.104139},
  url = {https://www.sciencedirect.com/science/article/pii/S0306457325000767}
}

@article{wang2026la2hdetect,
  title = {``Language Is the Dress of Thought'': A New Method for Automatic Detection of {AI}-Generated Text},
  author = {Wang, Zhenhua and Xu, Guang and Ren, Ming},
  journal = {Decision Support Systems},
  volume = {201},
  pages = {114578},
  year = {2026},
  issn = {0167-9236},
  doi = {10.1016/j.dss.2025.114578},
  url = {https://www.sciencedirect.com/science/article/pii/S0167923625001794}
}

@inproceedings{zhou2025adadetectgpt,
title = "AdaDetectGPT: Adaptive Detection of LLM-Generated Text with Statistical Guarantees",
author = "Hongyi Zhou and Jin Zhu and Pingfan Su and Kai Ye and Ying Yang and Shakeel Gavioli-Akilagun and Chengchun Shi",
year = "2025",
month = sep,
day = "17",
language = "English",
series = "Advances in Neural Information Processing Systems",
publisher = "NeurIPS",
booktitle = "Advances in Neural Information Processing Systems 38 (NeurIPS 2025)",
note = "The Thirty-Ninth Annual Conference on Neural Information Processing Systems<br/>, NeurIPS 2025 ; Conference date: 02-12-2025 Through 07-12-2025",
url = "https://neurips.cc/",
}

@inproceedings{li2024mage,
title = {{MAGE}: Machine-Generated Text Detection in the Wild},
author = {Li, Yafu and Li, Qintong and Cui, Leyang and Bi, Wei and Wang, Zhilin and Wang, Longyue and Yang, Linyi and Shi, Shuming and Zhang, Yue},
booktitle = {Proceedings of the 62nd Annual Meeting of the Association for Computational Linguistics (Volume 1: Long Papers)},
pages = {36--53},
year = {2024},
month = aug,
address = {Bangkok, Thailand},
publisher = {Association for Computational Linguistics},
doi = {10.18653/v1/2024.acl-long.3},
url = {https://aclanthology.org/2024.acl-long.3/}
}

@inproceedings{wang2024m4,
title = {{M4}: Multi-Generator, Multi-Domain, and Multi-Lingual Black-Box Machine-Generated Text Detection},
author = {Wang, Yuxia and Mansurov, Jonibek and Ivanov, Petar and Su, Jinyan and Shelmanov, Artem and Tsvigun, Akim and Whitehouse, Chenxi and Mohammed Afzal, Osama and Mahmoud, Tarek and Sasaki, Toru and Arnold, Thomas and Aji, Alham and Habash, Nizar and Gurevych, Iryna and Nakov, Preslav},
booktitle = {Proceedings of the 18th Conference of the European Chapter of the Association for Computational Linguistics (Volume 1: Long Papers)},
pages = {1369--1407},
year = {2024},
month = mar,
address = {St. Julian's, Malta},
publisher = {Association for Computational Linguistics},
url = {https://aclanthology.org/2024.eacl-long.83/}
}

@inproceedings{verma2024ghostbuster,
title = {Ghostbuster: Detecting Text Ghostwritten by Large Language Models},
author = {Verma, Vivek and Fleisig, Eve and Tomlin, Nicholas and Klein, Dan},
booktitle = {Proceedings of the 2024 Conference of the North American Chapter of the Association for Computational Linguistics: Human Language Technologies (Volume 1: Long Papers)},
pages = {1702--1717},
year = {2024},
month = jun,
address = {Mexico City, Mexico},
publisher = {Association for Computational Linguistics},
doi = {10.18653/v1/2024.naacl-long.95},
url = {https://aclanthology.org/2024.naacl-long.95/}
}

@inproceedings{ma2026nts,
title = {Zero-Shot Detection of {LLM}-Generated Text Using Temperature Sensitivity},
author = {Ma, Shixuan and Li, Jiahao and Mao, Zhendong and Wang, Quan},
booktitle = {Proceedings of the 64th Annual Meeting of the Association for Computational Linguistics (Volume 1: Long Papers)},
pages = {37664--37679},
year = {2026},
month = jul,
address = {San Diego, California, United States},
publisher = {Association for Computational Linguistics},
doi = {10.18653/v1/2026.acl-long.1748},
url = {https://aclanthology.org/2026.acl-long.1748/}
}

@article{guo2023chatgpt,
title = {How Close Is {ChatGPT} to Human Experts? Comparison Corpus, Evaluation, and Detection},
author = {Guo, Biyang and Zhang, Xin and Wang, Ziyuan and Jiang, Minqi and Nie, Jinran and Ding, Yuxuan and Yue, Jianwei and Wu, Yupeng},
journal = {arXiv preprint arXiv:2301.07597},
year = {2023},
eprint = {2301.07597},
archivePrefix = {arXiv},
primaryClass = {cs.CL},
doi = {10.48550/arXiv.2301.07597},
url = {https://arxiv.org/abs/2301.07597}
}

@inproceedings{yang2024dnagpt,
title = {{DNA-GPT}: Divergent {N}-Gram Analysis for Training-Free Detection of {GPT}-Generated Text},
author = {Yang, Xianjun and Cheng, Wei and Wu, Yue and Petzold, Linda and Wang, William and Chen, Haifeng},
booktitle = {International Conference on Learning Representations},
volume = {2024},
pages = {48572--48597},
year = {2024}
}

@inproceedings{xu2025training,
title = {Training-Free {LLM}-Generated Text Detection by Mining Token Probability Sequences},
author = {Xu, Yihuai and Wang, Yongwei and Bi, Yifei and Cao, Huangsen and Lin, Zhouhan and Zhao, Yu and Wu, Fei},
booktitle = {International Conference on Learning Representations},
volume = {2025},
pages = {19072--19098},
year = {2025}
}

@inproceedings{liu2026wavedetect,
title = {{WaveDetect}: Robust Framework for Machine-Generated Text Detection via Wavelet Transform},
author = {Liu, Zhichen and Qin, Kaitong and He, Linhan and Xu, Yang},
booktitle = {Findings of the Association for Computational Linguistics: ACL 2026},
pages = {8712--8727},
year = {2026},
month = jul,
address = {San Diego, California, United States},
publisher = {Association for Computational Linguistics},
isbn = {979-8-89176-395-1},
doi = {10.18653/v1/2026.findings-acl.424},
url = {https://aclanthology.org/2026.findings-acl.424/}
}

\end{document}